\documentclass{bmvc2k}
\usepackage{multirow}
\usepackage{amsmath}
\usepackage{amssymb}
\hypersetup{
  colorlinks   = true, 
  urlcolor     = blue, 
  linkcolor    = blue, 
  citecolor   = red, 
  linkbordercolor = red,
}

\title{Open-Vocabulary Temporal Action Detection with Off-the-Shelf Image-Text Features}

\addauthor{Vivek Rathod}{rathodv@google.com}{1}
\addauthor{Bryan Seybold}{seybold@google.com}{1}
\addauthor{Sudheendra Vijayanarasimhan}{svnaras@google.com}{1}
\addauthor{Austin Myers}{aom@google.com}{1}
\addauthor{Xiuye Gu}{xiuyegu@google.com}{1}
\addauthor{Vighnesh Birodkar}{vighneshb@google.com}{1}
\addauthor{David A. Ross}{dross@google.com}{1}

\addinstitution{
 Google Research\\
 Mountain View, CA\\
 USA
}
\runninghead{Rathod et al}{Open-Vocabulary Temporal Action Detection}

\AtBeginDocument{\hypersetup{pdfborder={0 0 0}}}

\begin{document}

\maketitle
\begin{abstract}
Detecting actions in untrimmed videos should not be limited to a small, closed set of classes. We present a simple, yet effective strategy for open-vocabulary temporal action detection utilizing pretrained image-text co-embeddings. Despite being trained on static images rather than videos, we show that image-text co-embeddings enable open-vocabulary performance competitive with fully-supervised models. We show that the performance can be further improved by ensembling the image-text features with features encoding local motion, like optical flow based features, or other modalities, like audio. In addition, we propose a more reasonable open-vocabulary evaluation setting for the ActivityNet data set, where the category splits are based on similarity rather than random assignment.
\end{abstract}

\section{Introduction}
\label{sec:intro}
Humans can learn to recognize new objects from only a few examples and/or simple descriptions because they can apply prior experience to do so. 
Recent Image-Text Co-Embedding (\textbf{ITCE}) models, such as CLIP~\cite{CLIP}, Align~\cite{Align}, and Basic~\cite{Basic}, demonstrate one path to achieve this goal in computer vision. These models learn a joint representation for images and texts using contrastive learning on Internet-scale image-text pairs and are able to capture the rich information present in text descriptions such as objects, actions, human-object interactions, and object-object relationships. ITCE features have demonstrated excellent open-vocabulary and zero-shot transfer performance on various image classification tasks without the need for additional fine-tuning or manually-annotated datasets. Recent works apply ITCE features to downstream open-vocabulary tasks such as object detection~\cite{OVRCNN,VILD}, action recognition~\cite{ActionClip,Clip4clip,EfficientPrompt} and---relatively underexplored---temporal action detection~\cite{EfficientPrompt}.

In this paper, we consider open-vocabulary Temporal Action Detection (\textbf{TAD}). We show that open-vocabulary TAD models using ITCE features outperform fully-supervised models from a year ago and approach the performance of concurrent work. TAD requires detecting and classifying actions in untrimmed videos. TAD is usually done in a fully-supervised setting: the training set contains videos annotated with the 3-tuples of segment start, end, and action labels. In the open-vocabulary setting, segments with a subset of action labels for evaluation are held out and unseen during the training phase then predicted during evaluation. We decouple the problem into learning to detect class-agnostic temporal segments and classifying the detected segments. Classification of video segments can be remarkably effective and straightforward using ITCE by simply averaging image features within each detected segment and comparing to the text embedding of each class label. We demonstrate open-vocabulary performance that is comparable to fully-supervised TAD and outperforms open-vocabulary models from the literature when utilizing strong image-text co-embedding features, such as from CLIP~\cite{CLIP}, Align~\cite{Align} and BASIC~\cite{Basic}. 

To summarize, our main contributions are:
\begin{itemize}
\setlength{\itemsep}{3pt}
\setlength{\parskip}{0pt}
\item A method on open-vocabulary temporal action detection that outperforms a large number of fully-supervised models.
\item Detailed evaluation of different image-text co-embedding features for both class agnostic action detection and open-vocabulary action classification.
\item A survey of features to complement image-text co-embedding features for class-agnostic detection to identify synergistic features that improve open-vocabulary performance.
\item A rationally designed ``Smart split'' for open-vocabulary temporal action detection on ActivityNet that leverages the class hierarchy for more meaningful evaluation.
\end{itemize}

\section{Related Work}
\paragraph{Open-vocabulary image-text co-embedding models and object detection.}
Pre-training on large manually-annotated classification datasets such as Imagenet~\cite{Imagenet} or JFT~\cite{JFT} is effective for learning image features that generalize to downstream tasks with limited training data, but manual annotation limits the size of datasets for training. Recent work on ITCEs such as CLIP~\cite{CLIP}, Align~\cite{Align} and BASIC~\cite{Basic} instead use contrastive training. The joint embedding spaces these models learn from weakly-labeled, Internet-scale datasets enabled downstream open-vocabulary classification tasks at high accuracy without requiring fine-tuning or manual annotation.

Open-vocabulary object detection aims to detect objects of unseen classes by learning to generalize from annotated objects~\cite{zeroshotod1,zeroshotod2,zeroshotod3}, textual descriptions~\cite{zeroshotod4} or captions~\cite{OVRCNN}. 
Recent work, such as \cite{VILD,OVRCNN}, decoupled the problem into learning class-agnostic proposal generators followed by open-vocabulary image classification using ITCE features and brought performance closer to fully-supervised detection than previous methods. 

\paragraph{Image-text co-embedding models for video and temporal action detection.}
Several recent arXiv papers\cite{ActionClip,Clip4clip,EfficientPrompt} leverage pre-trained CLIP~\cite{CLIP} features for video retrieval, action recognition and action localization. While \cite{ActionClip,Clip4clip} fine-tune the CLIP~\cite{CLIP} model end-to-end for video tasks, \cite{EfficientPrompt} focuses on efficiently adapting image features to videos tasks through temporal aggregation and gradient-based prompt-engineering.

TAD consists of detecting and classifying actions from untrimmed videos. TAD approaches can be roughly categorized as weakly-supervised methods that only train on video class labels~\cite{narayan2021d2net,islam2020weakly,UntrimmedNets}, self-contained methods~\cite{TalNet,SSN,CDC,PGCN,Tadtr} that are trained end-to-end to predict and classify segments and two-stage methods~\cite{GTAD,BMN,ActionFormer,rcl} that use various approaches to predict candidate segments and use state-of-the-art action classifiers~\cite{ANet2016Classifier} to classify them. However, all these approaches use the same fixed vocabulary of actions for both training and testing and are fully supervised. 

A few recent work attempt open-set TAD \cite{opental} and open-vocabulary TAD \cite{EfficientPrompt}. In \cite{opental}, the authors introduce the open-set TAL problem where a single 'unknown' label is introduced to model missing classes. Both our model and~\cite{EfficientPrompt} consider the more general open-vocabulary TAD problem and use two-stage models that replace the supervised classifier with a ITCE feature comparison. Our work differs from \cite{EfficientPrompt} along several axes: we compare ITCE features, we test features that can complement ITCE features, we explore more dataset splits, and we do not do prompt engineering. Through extensive experiments, we outperform other open-vocabulary TAD models~\cite{EfficientPrompt} by a large margin and approach the performance of fully supervised models from recent literature by using strong ITCE features.

\begin{figure}
\centering
\includegraphics[width=0.98\textwidth]{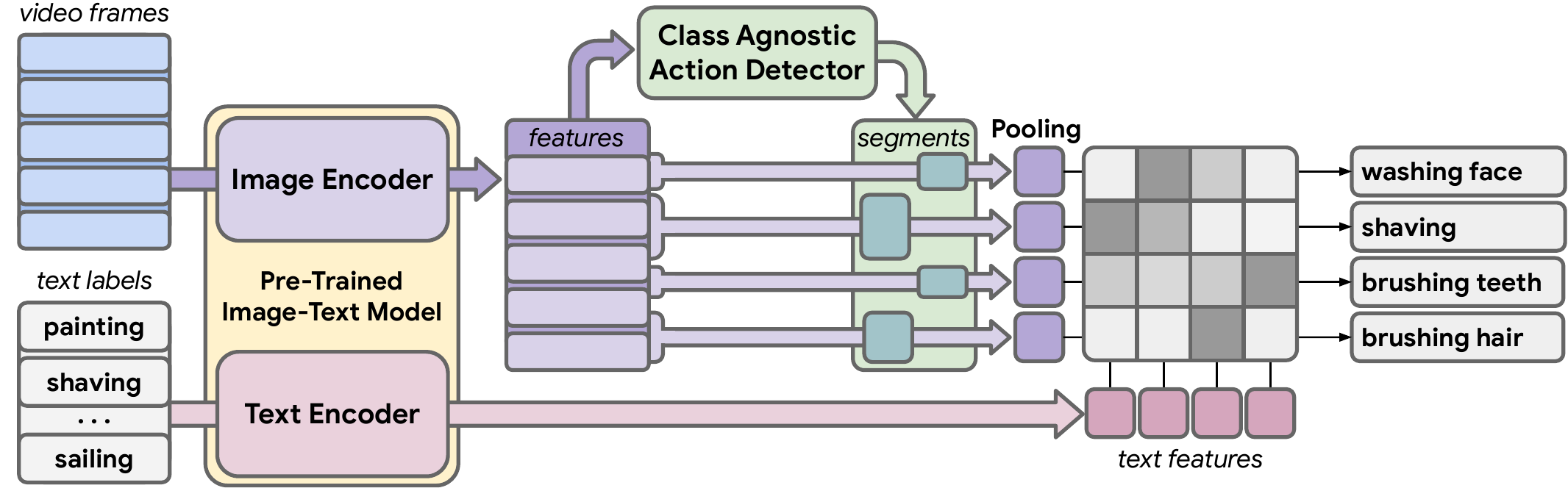}
\caption{
Overview of Open-Vocabulary Temporal Action Detection.
Video frames are processed by a pre-trained image encoder. These features or others are passed to a class-agnostic detector to detect temporal segments. Within each segment, the image features are pooled to form a segment feature. Finally, the segment feature is compared to the text feature of each class label of interest passed through a text encoder. The class labels during inference may be entirely novel.
}
\label{fig:architecture}
\end{figure}

\section{Method}

\subsection{Class-Agnostic Detection and Open-Vocabulary Classification}


To detect temporal actions, we first produce class-agnostic segment detections in each video and follow-up with open-vocabulary classification  as shown in Figure~\ref{fig:architecture}. We begin by extracting detection features, $f_{t}^{d} = F^{d}(X_t)$, and open-vocabulary features, $f^{ov}_{t} = F^{ov}(X_t)$, from video frames, $X_{1..T}$, where $F^{d}$ and $F^{ov}$ are pre-trained models. Both may be the same model; $F^{ov}$ is always an ITCE model; and both are applied with temporal windowing to produce one output feature per second of video. Following common TAD practices~\cite{TalNet,SSN,CDC,PGCN,Tadtr,GTAD,BMN}, we detect segments from the pre-extracted features using an architecture similar to object detection, $ S = D(f^{d}_{1..T}) $. To classify each segment, we pool the ITCE features within each segment along each dimension, $ f^{ov}_{S} = \mathop{\mathbb{E}}_{t \in S} [ f^{ov}_{t} ] $, compute the ICTE features for the text labels, $f^{ov}_{Y} = F^{ov}(Y) $ where $Y$ is the text of a class label, and then assign a class label that maximizes the softmax of the dot-product between the segment feature and text feature: $ arg\,max(softmax( f^{ov}_{S} \cdot f^{ov}_{Y})) $.


As open-vocabulary TAD is a new area, we focus on a solid foundation evaluating off-the-shelf models and features. The novelty of our work is that ITCE features enable straightforward methods to outperform other open-vocabulary work and nearly match state-of-the-art, fully-supervised results without needing complicated tricks. We focus on using out-of-the-box detection models that achieve good performance on temporal action detection rather than relying on models that may be over-tuned to use specific input features and datasets common in the literature. More complicated models for open-vocabulary action detection should improve our results, but we leave that for future research because including more complicated models would make it difficult to communicate how effective simple models are with ITCE features.



\subsection{Smart Split}\label{sec:rational_split}

\begin{table}[h]
    \centering
    \scalebox{0.8}{
    \begin{tabular}{|c|c|c|c|}
    \hline
    \multicolumn{4}{|c|}{Smart Split} \\
    \hline

Shot put & Hammer throw & Long jump & Smoking a cigarette\\
Spinning & High jump & Vacuuming floor & Using the pommel horse\\
Camel ride & Playing racquetball & Hurling & Playing ice hockey\\
Kayaking & Wakeboarding & Rafting & Longboarding\\
Assembling bicycle & Zumba & Belly dance & Painting furniture\\
Playing accordion & Playing flauta & Playing violin & Tennis serve with ball bouncing\\
Skiing & Playing congas & Drum corps & Futsal\\
Playing beach volleyball & Doing karate & Playing badminton & Getting a haircut\\
Playing kickball & Doing motocross & Cutting the grass & Making an omelette\\
Preparing salad & Grooming horse & Washing face & Blowing leaves\\
Shaving legs & Plataform diving & Polishing shoes & Mixing drinks\\
Painting fence & Roof shingle removal & Clipping cat claws & Windsurfing\\
River tubing & Waterskiing &  & \\

\hline
    \end{tabular}
    }
    \caption{ActivityNet Smart 75/25 Split.
     The strings listed are the ActivityNet labels held out for evaluation. This
     split is designed so that the classes in evaluation are similar to classes
     in the training set.}
    \label{tab:anet_smart_split_supp}
\end{table}

Open-vocabulary detection is commonly evaluated in the literature by separating different classes of a well-known data set into new training and evaluation splits~\cite{EfficientPrompt,VILD,OVRCNN}. 
One simple way to split the data is to randomly select a subset of all classes to hold out. Random splits artificially limit open-vocabulary performance compared to fully-supervised methods because it breaks the distribution match training and evaluation sets. In contrast, fully-supervised methods are generally evaluated with matching distributions. 

Even defining action boundaries in time is notoriously difficult, so care is always needed to ensure TAD evaluations are meaningful. The two most common temporal action detection data sets, ActivityNet \cite{ActivityNet} and Thumos \cite{Thumos} disagree on the length of actions with even the same label by an order of magnitude, and data set creators need to defend whether the label annotators are able to be consistent \cite{alwassel_2018_detad}. With a random split, many errors will simply occur because the evaluation set is out of domain for the model, which are uninteresting errors.

To address the mismatch, we propose a new split of the ActivityNet dataset that is designed so that classes in evaluation are similar to classes in the training set. Table~\ref{tab:anet_smart_split_supp} shows the 25\% of the classes selected to be in the evaluation split with the remaining 75\% of the classes used for training. We do this to ensure that for each class in the evaluation set, some class in the training set has similar boundaries by taking neighbors from the AcitivityNet class ontology. 
We hope that this split is useful to the TAD community for evaluating open-vocabulary TAD going forward.. 

\subsection{Complementary Features}

ITCE features based on images will not perfectly encode video because they operate on each frame individually.
One possible improvement is to adapt ITCEs to take multiple frames into account \cite{ActionClip,Clip4clip,EfficientPrompt}, but these methods risk overfitting and have not yet been proven to reliably encode temporal information. In contrast, we include time-tested features such as I3D\_RGB and I3D\_FLOW \cite{I3D} that have been shown to encode temporal information, or explicit audio features \cite{Yamnet,Vggish}. These features can establish a meaningful baseline for evaluating future work to produce new video features.

\section{Experiments}

\subsection{Data sets and splits}

\textbf{Data sets:} We focus on the two most common temporal action detection data sets in the literature: ActivityNet 1.3 \cite{ActivityNet} and Thumos14 \cite{Thumos}. We use the entire Thumos data set and the ~90\% of ActivityNet that is still accessible. We verify that our comparable models using CLIP B/16 embeddings achieve comparable performance to a recent paper using the same features \cite{EfficientPrompt} (see results) despite any change in the data.

\noindent\textbf{Random splits:} For each data set, we create two sets of random splits: 12 random splits with 75\% of the data in the training set and 25\% held out for evaluation; as well as another 4 random splits with 50\% in the training set and 50\% held out for evaluation. The exact labels used for the evaluation set of each split are provided in the supplemental. (We compared class-agnostic detection performance on our splits versus the 10 random splits from \cite{EfficientPrompt} and found them virtually identical. See supplemental.) 
Unless otherwise noted, we present numbers as mean $\pm$ standard error of the mean when presenting results on random splits. If one video has labels from both the training and evaluation splits, the same video appears in both splits, but with only the labels in that split. This configuration is challenging because true-positives for the training set hurt performance during evaluation, but accurately maps onto real uses of open-vocabulary detection.

\noindent\textbf{Smart split:} As described in \ref{sec:rational_split}, intelligently selecting an open-vocabulary split with related concepts in the training and evaluation sets is a more meaningful evaluation than mismatched concepts. We propose the \textbf{Smart split} for ActivityNet based on the provided class hierarchy with 25\% of the labels in the evaluation set. We created the Smart split by taking pairs of neighboring leaf nodes from the hierarchy and ensuring that one was held out for evaluation and the other was used for training. Based on a subjective assessment, we only chose pairs that were perceptually similar. (E.g. ``Playing rubik cube'' is visually distinct from all other classes, and would not be used in a pair.) We provide the exact labels used for the evaluation set of the Smart split in the supplemental and hope future work will report on this split to give a meaningful comparison without needing to test as many split combinations.

We do not make a Smart split for Thumos as there is no class hierarchy and the data set contains only 20 classes.

\subsection{Models} 

We use two different models on the two different data sets because we find that different out-of-the-box detection models perform much better on the fully supervised temporal action detection task (see supplemental for model details). 

For ActivityNet, we use a DeTr \cite{DETR} detection head on top of a stacked-hourglass \cite{newell2016stacked}-style feature extractor. The feature extractor stacks 1D convolution residual blocks with LayerNorm \cite{LayerNorm} and ReLU activations in a U-Net pattern of downsampling and upsampling. Additional residual connections are passed between blocks at the same resolution to avoid loss of temporal information. Four residual U-Net stacks are applied in turn before the detection head. For Thumos, we use a CenterNet \cite{CenterNet} detection head on top of the same style of feature extractor. Both detection heads are adapted for 1D detection by treating all proposals and ground truth segments as boxes with a fixed height of one. The CenterNet code is adapted from \cite{ODapi}. On the fully-supervised TAL task, these models achieve competitive performance each data set using TSP \cite{TSP} features: ActivityNet, 28\% mAP Avg @ IoU 0.5:0.95:0.05 (without using external, whole-video classification); Thumos, 52\% mAP @ IoU 0.5. The TSP features are trained on all 200 classes of ActivityNet and all 20 of Thumos, so we cannot evaluate TSP features for open-vocabulary performance. 

Key hyperparameters for the models and optimization: DeTr: transformer heads is set to 16, transformer dimensions are 1024, segment classifier's hidden layer size is 1024, maximum number of proposals is set to 64, background class weight is set to 3.0, weight decay is set to 0.0005 and the positional encoding is learned. CenterNet: maximum number of segments is set to 512, scale loss weight is 0.1, heatmap head dimensions is set to 256. Residual Feature Extractor: base hidden dimensions is set to 512, resolutions is set to 5, number of stacked U-Nets is set to 4, kernel size is set to 5. We use the Adam optimizer with a learning rate of 0.0001 and train for 20,000 steps. See supplemental for additional details.




\subsection{Features}
We use the following features: I3D\_RGB and I3D\_FLOW \cite{I3D} features trained on Kinetics \cite{Kinetics}; VIVIT \cite{Vivit} features trained on Kinetics; VGGISH \cite{Vggish} features trained on AudioSet \cite{AudioSet}; YAMNET \cite{Yamnet} features pre-trained on AudioSet; ALIGN \cite{Align} features trained on the Align data set; BASIC \cite{Basic} features trained on Align+JFT \cite{JFT} data sets; MBT-a \cite{nagrani2021attention} trained on AudioSet \cite{AudioSet}; MBT-v \cite{nagrani2021attention} trained on Kinetics \cite{Kinetics}; and CLIP \cite{CLIP} B/16, B/32, and L/14 features trained on WebImageText. We extract each feature at 1 FPS on both data sets using a sliding window for features that take more than 1 second of input.

We do not use video-text co-embeddings because the performance of related models was worse than image-text models without finetuning (see supplemental and literature ~\cite{nagrani2022learningvideocc}). Most video-text models are still fine-tuned for specific tasks due a lack of large scale video-text datasets~\cite{ActionClip,Clip4clip}, and we evaluate other fine-tuned video transformer features~\cite{Vivit,MBT}.

\subsection{ITCE Features Empower Open-Vocabulary Action Classifiers}

To motivate this work, we first test how powerful ITCE features are for ground truth segment classification. For each ground truth segment, we pool the 1 FPS features from different models then assign the label from the closest text embedding.
In general, the performance on ground truth segments in Table~\ref{tab:gt_classification} scales with model size for the original splits and the Smart split. The Smart split is as difficult to classify as the original data. The performance of the BASIC image-text co-embeddings exceeds the Top-1 accuracy of fully-supervised ActivityNet video classification models like \cite{ANet2016Classifier} commonly used to do classification on top of temporal action proposal networks in recent literature.
Using off-the-shelf ITCE features enable better classification than using the ActivityNet training data!

\begin{table}[t]
    \centering
    \scalebox{0.75}{
    \begin{tabular}{|c|c|c|c|c|c|c|c|c|}
    \hline
      \multirow{3}{*}{Features} & \multirow{3}{*}{Setup} & \multicolumn{3}{c|}{ActivityNet} & \multicolumn{2}{c|}{Thumos} & \multicolumn{2}{c|}{\#params} \\
      \cline{3-9}
      & & \multirow{2}{*}{Top-1} &\multirow{2}{*}{Top-5} & Smart split & \multirow{2}{*}{Top-1} & \multirow{2}{*}{Top-5} & \multirow{2}{*}{Image tower} & \multirow{2}{*}{Text tower}\\
      & &  &  & Top-1 &  &  &  &  \\
      \hline
2016 ANet winner \cite{ANet2016Classifier} & Supervised & 88.1 & - & - & - & - & - & -\\
CLIP ViT-B/32 \cite{CLIP} & Open-Vocab & 65.38 & 89.26 & 65.95 & 61.61 & 90.11 & 87 M & 63 M\\
CLIP ViT-B/16 \cite{CLIP} & Open-Vocab & 71.29 & 92.41 & 66.32 & 64.26 & 93.58 & 87 M & 63 M\\
CLIP ViT-L/14 \cite{CLIP} & Open-Vocab & 78.2 & 95.74 & 75.38 & 70.59 & 97.05 & 303 M & 123 M\\
ALIGN \cite{Align} & Open-Vocab & 77.92 & 94.53 & 76.06 & 42.83 & 70.35 & 632 M & 632 M\\
BASIC \cite{Basic} & Open-Vocab & \textbf{88.73} & \textbf{97.52} & \textbf{88.29} & \textbf{81.86} & \textbf{98.2} & 2.4 B & 670 M \\ 
\hline
    \end{tabular}}
  \vspace{0.1in}
    \caption{Open-Vocabulary Classification of Ground Truth Segments. We present Top-1 and Top-5 classification accuracies on ground truth segments by comparing pooled features to text label embeddings for ActivityNet (original 200-class split and 50-class Smart split) and Thumos (the original 20-class split). We also list the number of parameters in the image and text towers of each image-text model.}
    \label{tab:gt_classification}
\end{table}

\subsection{Class-Agnostic Detection}

We evaluate our models for class-agnostic detection on random open-vocabulary splits of each data set to match the common open-vocabulary paradigm and the Smart split of ActivityNet. We present both an average of mean Average Precision (mAP) at different Intersection over Unions (IoUs) and average recall at N metrics for comparison to the literature, but emphasize that AP metrics are more appropriate: For open-vocabulary detection, the classifier cannot classify segments as background so these are detections, rather than merely proposals. Precision matters as much as recall for detections.

\begin{table}[t]
    \centering
    \scalebox{0.58}{
    \begin{tabular}{|c|c|c|c|c|c|c|c|c|c|c|}
    \hline
  & \multicolumn{6}{c|}{ActivityNet} & \multicolumn{4}{c|}{Thumos} \\
 \cline{2-11}
  \multirow{3}{*}{Features} &  \multicolumn{2}{c|}{75/25 splits} & \multicolumn{2}{c|}{Smart split}  & \multicolumn{2}{c|}{50/50 splits}  &  \multicolumn{2}{c|}{75/25 splits}  & \multicolumn{2}{c|}{50/50 splits} \\
   & \multicolumn{2}{c|}{average (n=12)}   &  \multicolumn{2}{c|}{}  & \multicolumn{2}{c|}{average (n=4)} & \multicolumn{2}{c|}{average (n=12)}  & \multicolumn{2}{c|}{average (n=4)} \\
 \cline{2-11}
 & AP@avg & AR@10 & AP@avg & AR@10 & AP@avg & AR@10 & AP@avg & AR@10 & AP@avg & AR@10\\
\hline
CLIP b/32 \cite{CLIP} & 27.2 $\pm$ 1.0 & 41.3 $\pm$ 0.9 & 29.2 & 43.0 & 25.6 $\pm$ 0.4 & 38.6 $\pm$ 0.9 & 32.2 $\pm$ 7.7 & 14.3 $\pm$ 3.2 & 16.4 $\pm$ 4.5 & 8.1 $\pm$ 2.6 \\
CLIP b/16 \cite{CLIP} & 27.6 $\pm$ 1.0 & 42.4 $\pm$ 1.2 & 30.7 & 44.0 & 25.8 $\pm$ 0.7 & 40.0 $\pm$ 1.3 & 32.1 $\pm$ 8.0 & 14.1 $\pm$ 3.5 & 20.4 $\pm$ 4.0 & 9.8 $\pm$ 1.9 \\
CLIP l/14 \cite{CLIP} & 28.2 $\pm$ 0.8 & 42.5 $\pm$ 0.9 & 31.3 & 43.4 & 26.6 $\pm$ 0.6 & 40.1 $\pm$ 0.7 & 34.8 $\pm$ 7.9 & 15.6 $\pm$ 3.2 & 18.3 $\pm$ 5.7 & 8.7 $\pm$ 2.0 \\
ALIGN \cite{Align} & 27.8 $\pm$ 0.9 & 42.3 $\pm$ 0.8 & 30.3 & 43.0 & 26.1 $\pm$ 0.9 & 40.1 $\pm$ 1.0 & 31.7 $\pm$ 4.9 & 14.2 $\pm$ 3.3 & 20.9 $\pm$ 4.2 & 9.7 $\pm$ 2.3 \\
BASIC \cite{Basic} & \textbf{28.8 $\pm$ 0.9} & \textbf{43.3 $\pm$ 0.8} & \textbf{32.0} & \textbf{44.9} & \textbf{27.0 $\pm$ 0.8} & \textbf{41.0 $\pm$ 1.1} & 33.7 $\pm$ 7.9 & 15.2 $\pm$ 3.5 & 20.2 $\pm$ 2.8 & 10.4 $\pm$ 2.2 \\
I3D RGB+FLOW & 28.4 $\pm$ 0.9 & 42.3 $\pm$ 0.8 & 30.4 & 43.1 & 26.5 $\pm$ 0.5 & 40.4 $\pm$ 0.7  & \textbf{38.6 $\pm$ 9.3} & \textbf{16.1 $\pm$ 3.9} & \textbf{26.9 $\pm$ 6.9} & \textbf{11.7 $\pm$ 3.0}\\
I3D RGB \cite{I3D} & 27.2 $\pm$ 0.7 & 41.3 $\pm$ 0.7 & 30.7 & 43.8 & 25.1 $\pm$ 0.3 & 38.9 $\pm$ 0.3 &&&&\\
I3D FLOW \cite{I3D} & 28.3 $\pm$ 1.0 & 42.2 $\pm$ 1.0 & 29.6 & 42.5 & 25.6 $\pm$ 0.5 & 40.3 $\pm$ 01.3 &&&&\\
YAMNET \cite{Yamnet} & 22.7 $\pm$ 0.6 & 36.6 $\pm$ 0.6 & 25.5 & 40.1 & 20.9 $\pm$ 1.0 & 34.9 $\pm$ 01.7 &&&&\\
VGGISH \cite{Vggish} & 23.0 $\pm$ 0.8 & 36.9 $\pm$ 0.9 & 25.0 & 38.8 & 21.3 $\pm$ 0.5 & 35.6 $\pm$ 1.1 &&&&\\
VIVIT \cite{Vivit}  & 22.4 $\pm$ 0.8 & 37.1 $\pm$ 1.1 & 24.8 & 40.0 & 20.0 $\pm$ 0.4 & 34.2 $\pm$ 1.1 &&&&\\
MBT-v \cite{MBT} & 23.5 $\pm$ 0.7 & 38.5 $\pm$ 0.8 & 25.6 & 40.8 & 21.8 $\pm$ 0.4 & 36.9 $\pm$ 0.9 &&&&\\
MBT-a \cite{MBT} & 18.1 $\pm$ 0.8 & 34.1 $\pm$ 1.0 & 20.0 & 35.1 & 17.9 $\pm$ 0.3 & 33.6 $\pm$ 0.5 &&&&\\
\hline
    \end{tabular}}
    \vspace{0.1in}
    \caption{Class-Agnostic Detection Results. Performance of the same class-agnostic detector architectures with different input features. On ActivityNet, mAP@Avg is mAP@0.5:0.05:0.95. On Thumos, mAP@avg is mAP@0.3:0.1:0.7. AR@10 is Average Recall at 10. Data is grouped as columns for 75/25 random splits (n=12) and Smart split, and 50/50 random splits (n=4). Aggregates are means $\pm$ standard error across splits. Some features were not tested on Thumos. See supplemental for additional metrics.}
    \label{tab:anet_proposals}
\end{table}

In Table \ref{tab:anet_proposals}, we find that ITCE features are effective for class-agnostic temporal detection. (These can be compared to fully-supervised temporal action detection in Table~\ref{tab:anet_e2e} and supplemental). As with ground truth classification performance, ITCE features from larger models perform better. The I3D\_RGB $+$ I3D\_FLOW features commonly used in the TAD literature are also very good for class-agnostic detection. Audio features alone do surprisingly well, and recent video transformers trained for classification such as ViViT and MBT do surprisingly poorly. (In the supplemental, we show that doing fully supervised training with classification restores the expected order that modern ViViT and MBT do well while audio features alone are insufficient to classify visual actions.)
The Smart split is typically 1-2\% AP better than the average of random splits. This difference highlights that a major challenge for open-vocabulary temporal action detection is a difficulty recognizing segments of novel classes. 
Because Thumos contains so few classes, evaluation on the held-out evaluation classes jumps by large amounts whenever a new class is learned (see figure in supplemental). Because some classes are harder to learn than others depending on the splits, the performance across splits has much larger variability for Thumos than for ActivityNet. The standard error is 10-times larger: ~8\% mAP for Thumos and ~0.9\% mAP for ActivityNet. The adoption of larger data sets in future work will hopefully decrease the variance in evaluations.

\subsection{Open Vocabulary Temporal Action Detection}

We evaluate the open-vocabulary temporal action detection performance of each ITCE feature in Table \ref{tab:anet_e2e}. We evaluate each feature two ways. We use the ITCE feature for both detections and classification, and we also use the ITCE feature for classification on top of I3D\_RGB $+$ I3D\_FLOW detections. Using the same features and a similar architecture to BaselineIII from \cite{EfficientPrompt}, our CLIP-b/16  results are slightly better (21.4 $\pm$ 0.7 vs 20.2). They fine-tune their CLIP-b/16 features on ActivityNet and perhaps that causes overfitting. Using BASIC features results in a substantial improvement in performance that substantially outperforms~\cite{EfficientPrompt}.

\begin{table}[t]
    \centering
    \scalebox{0.7}{
    \begin{tabular}{|c|c|c|c|c|c|c|c|}
    \hline
       & \multicolumn{4}{c|}{ActivityNet} & \multicolumn{3}{c|}{Thumos} \\
       \cline{2-8}
 & \multicolumn{2}{c|}{75/25 Splits} & 50/50 Splits & fully-supervised & 75/25 Splits & 50/50 Splits & fully-supervised \\
Detector$\rightarrow$Classifier & mAP@avg & Smart & mAP@avg & mAP@avg & mAP@avg & mAP@avg & mAP@avg \\
& (n=12) &  & (n=4)& &(n=12) & (n=4) & \\
\hline
Clip b/32$\rightarrow$Clip b/32 & 19.4 $\pm$ 0.4 & 22.6 & 17.3 $\pm$ 0.6 & 21.5 & 12.0 $\pm$ 1.4 & 8.0 $\pm$ 1.8 & 26.6\\
Clip b/16$\rightarrow$Clip b/16 & 21.4 $\pm$ 0.7 & 23.5 & 19.5 $\pm$ 0.4 & 24.0 & 12.9 $\pm$ 1.3 & 9.1 $\pm$ 0.8 & 29.0\\
Clip l/14$\rightarrow$Clip l/14 & 24.6 $\pm$ 0.4 & 28.7 & 22.4 $\pm$ 0.7 & 26.1 & 15.8 $\pm$ 1.6 & 10.5 $\pm$ 1.9 & 32.6\\
Align$\rightarrow$Align & 24.2 $\pm$ 0.7 & 28.1 & 22.9 $\pm$ 1.1 & 25.9 & 8.7 $\pm$ 1.5 & 6.5 $\pm$ 1.4 & 32.7 \\
Basic$\rightarrow$Basic & \textbf{28.4 $\pm$ 0.5} & 31.2 & \textbf{26.4 $\pm$ 0.9} & \textbf{29.4} & 20.8 $\pm$ 2.3 & 16.5 $\pm$ 2.7 & \textbf{45.4}\\
I3D$\rightarrow$Clip b/32 & 20.2 $\pm$ 0.4 & 23.9 & 18.2 $\pm$ 0.6 &  & 14.1 $\pm$ 1.7 & 11.5 $\pm$ 2.6 &\\
I3D$\rightarrow$Clip b/16 & 22.1 $\pm$ 0.5 & 24.1 & 20.1 $\pm$ 0.5 &  & 15.6 $\pm$ 1.9 & 12.9 $\pm$ 3.1 &\\
I3D$\rightarrow$Clip l/14 & 24.8 $\pm$ 0.5 & 28.1 & 22.8 $\pm$ 0.6 &  & 17.0 $\pm$ 1.7 & 14.5 $\pm$ 3.2 &\\
I3D$\rightarrow$Align & 24.8 $\pm$ 0.6 & 28.4 & 22.9 $\pm$ 1.0 &  & 10.4 $\pm$ 1.5 & 8.3 $\pm$ 2.1 &\\
I3D$\rightarrow$Basic & 28.2 $\pm$ 0.5 & \textbf{31.3} & \textbf{26.4 $\pm$ 0.7} &  & \textbf{22.9 $\pm$ 2.1} & \textbf{19.5 $\pm$ 4.6} &\\
\hline
EP-baselineIII~\cite{EfficientPrompt} & 20.2 &  &  16 &  & 18.8 * & 15.7 * &\\
EP-full model~\cite{EfficientPrompt} & 23.1 &  & 19.6 &  & 23.3 * & 21.9 * &\\
\hline
    \end{tabular}}
    \vspace{0.1in}
    \caption{ Open-Vocabulary Temporal Action Detection. Evaluation when using a detector on one set of features and classifying those segments with the classifier to the $\rightarrow$. On ActivityNet, mAP@Avg is mAP@0.5:0.05:0.95. On Thumos, mAP@avg is mAP@0.3:0.1:0.7. The * on the bottom rows indicates evaluations where the test set videos were trimmed to evaluation classes only. See supplemental for additional metrics.}
    \label{tab:anet_e2e}
\end{table}

The Thumos data set is much more difficult. In our evaluation, which mimics expected use scenarios, true-positives for the training classes are false-positives during evaluation. This differs from the evaluation approach in \cite{EfficientPrompt} that splits each long video with multiple class labels, by an undescribed method, into short video clips with only one label and only evaluates on the subset of video clips with evaluation set labels.  Doing so leaks information about the location of the evaluation segments and removes likely false positives from the evaluation. We cannot compare directly against this undescribed method, but provide their numbers in Table \ref{tab:anet_e2e} for reference. Despite our more challenging evaluation procedure, we match the EP-baselineIII results and are within 2\% mAP avg on 50/50 splits and nearly match performance on 75/25 splits of the EP-full model.


\begin{table}[b!]
    \centering
    \scalebox{0.9}{
    \begin{tabular}{|c|c|c|c|c|c|c|}
    \multicolumn{7}{c}{\large ActivityNet - 75/25 split average (n=12)} \\
    \hline
     Features & CLIP-b/32 & CLIP-b/16 & CLIP-l/14 & ALIGN & BASIC & avg. diff. \\
     \hline
     B & 19.4 $\pm$ 0.4 & 21.4 $\pm$ 0.7 & 24.6 $\pm$ 0.4 & 24.2 $\pm$ 0.7 & 28.4 $\pm$ 0.5 & N/A\\
B+F & 20 $\pm$ 1 & 22.1 $\pm$ 0.9 & 24.8 $\pm$ 0.9 & 25.3 $\pm$ 1.3 & 28.6 $\pm$ 1 & 0.935\\
B+F+A & 20 $\pm$ 1 & 22 $\pm$ 0.9 & 24.8 $\pm$ 0.9 & 25.3 $\pm$ 1.4 & 28.3 $\pm$ 1 & 0.887\\
B+F+A+V & 20.2 $\pm$ 1 & 22.4 $\pm$ 1.1 & 25.1 $\pm$ 0.8 & 25.3 $\pm$ 1.3 & 28.8 $\pm$ 0.9 & 1.130\\
\hline
 \multicolumn{7}{c}{\large ActivityNet - Smart split} \\
 \hline
B & 22.6 & 23.5 & 25.7 & 26.9 & 29.5 & N/A\\
B+F & 23.5 & 23.9 & 27.2 & 28.7 & 32.2 & 1.5\\
B+F+A & 24.3 & 24.6 & 28.6 & 29.4 & 32.0 & 2.2\\
B+F+A+V & 23.8 & 24.4 & 28.2 & 29.3 & 32.1 & 1.9\\
\hline
  \multicolumn{7}{c}{\large Thumos - 75/25 split average (n=12)}  \\
 \hline
B & 12 $\pm$ 1.4 & 12.9 $\pm$ 1.3 & 15.8 $\pm$ 1.6 & 8.7 $\pm$ 1.5 & 20.8 $\pm$ 2.3 & N/A\\
B+F & 11.5 $\pm$ 3.3 & 12.8 $\pm$ 2.4 & 17.7 $\pm$ 2.4 & 9.4 $\pm$ 2.5 & 18.9 $\pm$ 4.6 & 0.102\\
B+F+A & 12.7 $\pm$ 3.1 & 14.9 $\pm$ 3.6 & 18.6 $\pm$ 2.6 & 7.2 $\pm$ 2.2 & 20.1 $\pm$ 4.2 & 0.769\\
B+F+A+V & 13.9 $\pm$ 3.2 & 16 $\pm$ 4 & 16.9 $\pm$ 3 & 10.4 $\pm$ 3 & 21.1 $\pm$ 4.3 & 1.725 \\
\hline
     \end{tabular}}
    \caption{Open Vocabulary Detection with Ensembled Detection Features. The benefits of ensembling synergistic features
    for open-vocabulary detection: B - the base ITCE feature alone; B+F - adding I3D\_FLOW; B+F+A - also adding VGGISH audio; B+F+A+V - also adding I3D\_RGB video features. ActivityNet split average is mAP@0.5:0.05:0.95. Thumos split average is mAP@0.3:0.1:0.7. See supplemental for additional metrics.}
    \label{tab:merged_feature_detections}
\end{table}

\subsection{What's Missing from Image-Text Co-Embedding Features?}

We study what information is missing from ITCE features by determining which features improve open-vocabulary TAD performance when concatenated with the ITCE features. In the supplemental, we present a large number of feature combinations on both class-agnostic and open-vocabulary detection, and only present the most synergistic combination in Table~\ref{tab:merged_feature_detections}.
Ensembling I3D\_RGB $+$ I3D\_FLOW $+$ VGGISH features with BASIC features is able to boost models above 32\% Avg AP@0.5:0.05:0.95 on the Smart split of ActivityNet. Across the board we often seem 1-2\% mAP avg improvements when ensembling features, although results on Thumos are noisy.
These particular features capture complementary information about short-term dynamics (e.g. I3D\_FLOW) or different modalities (e.g. VGGISH). 
The feature that produced the largest, reliable boost is I3D\_FLOW. 

\subsection{Comparison of Open-vocabulary Methods to State of the Art}


\begin{table}[t]
    \centering
    \scalebox{0.81}{
    \begin{tabular}{|c|c|c|c|c|c|c|c|c|c|c|}
    \hline
    \multicolumn{5}{|c|}{Open vocabulary} & \multicolumn{3}{c|}{Supervised, self-contained} & \multicolumn{3}{c|}{Supervised, ensemble} \\
    \hline
    \multicolumn{2}{|c|}{EP-full \cite{EfficientPrompt}} & \multicolumn{3}{c|}{Ours (best)} & P-GCN & PCG-TAL & TadTR & AFSD & Action & TadTR \\
    \cline{1-5}
    75/25 & 50/50 & 75/25 & 50/50  & Smart & \cite{PGCN} & \cite{su2020pcg} & \cite{Tadtr} & \cite{lin2021learningafsd} & Former\cite{ActionFormer}  &  \cite{Tadtr}\\
    \hline
    23.1 & 19.6 & \textbf{28.8} & \textbf{26.4} & \textbf{32.2} & 27.0 & 27.34 & \textbf{33.85} & 34.39 & 36.0 & 36.8 \\
    \hline    
    \end{tabular}}
    \vspace{0.1in}
    \caption{State-of-the-art comparison on ActivityNet across different methods. Numbers are mAP@0.5:0.05:0.95 performance on ActivityNet. ``Supervised, Self-contained'' methods may use pretraining and train on ActivityNet TAD, but do not use auxiliary classifiers trained on ActivityNet video classification which ``Supervised, ensemble'' methods do.}
    \label{tab:anet_sota}
\end{table}

We compare our model against those in the literature in Table~\ref{tab:anet_sota} with further comparisons in the supplemental. Ensembling BASIC features with I3D RGB+FLOW and Yamnet audio features results in impressive open-vocabulary performance of 28.8\% mAP avg on the random 75/25 splits and 32.2\% mAP avg on the Smart split. This performance compares well to self-contained fully-supervised models in the literature that do not use dedicated ActivityNet video classification models. It outperforms models from a year ago~\cite{PGCN,su2020pcg} and approaches the performance of concurrent work ~\cite{Tadtr}. Our model still trails ensemble models that use dedicated ActivityNet video classification models~\cite{Tadtr,ActionFormer,lin2021learningafsd}, but are much closer than other open-vocabulary work~\cite{EfficientPrompt}.

\section{Conclusion}

In conclusion, we show that ITCEs applied to open-vocabulary temporal action detection are very powerful and almost match the performance of fully supervised models of similar complexity. These features can be complemented by adding flow and audio based features that capture information about short-term dynamics or different modalities. Due to its high performance and relative simplicity, our approach proves the potential of open-vocabulary TAD to supplant fully-supervised methods, extending the capabilities of video action detection models to classes of events and users not well represented by existing fully-supervised datasets. From this basis, we hope future work will extend our straightforward open-vocabulary temporal action detection with models that incorporate more task-specific elements, such as prompt-engineering or bottom-up boundary prediction, and we hope for a broader range of data sets that define segment boundaries in consistent, meaningful ways.

\clearpage

\bibliography{egbib}

\begin{thebibliography}{53}
\providecommand{\natexlab}[1]{#1}
\providecommand{\url}[1]{\texttt{#1}}
\expandafter\ifx\csname urlstyle\endcsname\relax
  \providecommand{\doi}[1]{doi: #1}\else
  \providecommand{\doi}{doi: \begingroup \urlstyle{rm}\Url}\fi

\bibitem[Alwassel et~al.(2018)Alwassel, Caba~Heilbron, Escorcia, and
  Ghanem]{alwassel_2018_detad}
Humam Alwassel, Fabian Caba~Heilbron, Victor Escorcia, and Bernard Ghanem.
\newblock Diagnosing error in temporal action detectors.
\newblock In \emph{The European Conference on Computer Vision (ECCV)},
  September 2018.

\bibitem[Alwassel et~al.(2021)Alwassel, Giancola, and Ghanem]{TSP}
Humam Alwassel, Silvio Giancola, and Bernard Ghanem.
\newblock {TSP: Temporally-Sensitive Pretraining of Video Encoders for
  Localization Tasks}.
\newblock In \emph{Proceedings of the IEEE/CVF International Conference on
  Computer Vision (ICCV) Workshops}, 2021.

\bibitem[Arnab et~al.(2021)Arnab, Dehghani, Heigold, Sun, Lu{\v{c}}i{\'c}, and
  Schmid]{Vivit}
Anurag Arnab, Mostafa Dehghani, Georg Heigold, Chen Sun, Mario Lu{\v{c}}i{\'c},
  and Cordelia Schmid.
\newblock Vivit: A video vision transformer.
\newblock In \emph{Proceedings of the IEEE/CVF International Conference on
  Computer Vision}, pages 6836--6846, 2021.

\bibitem[Ba et~al.(2016)Ba, Kiros, and Hinton]{LayerNorm}
Jimmy~Lei Ba, Jamie~Ryan Kiros, and Geoffrey~E Hinton.
\newblock Layer normalization.
\newblock \emph{arXiv preprint arXiv:1607.06450}, 2016.

\bibitem[Bansal et~al.(2018)Bansal, Sikka, Sharma, Chellappa, and
  Divakaran]{zeroshotod1}
Ankan Bansal, Karan Sikka, Gaurav Sharma, Rama Chellappa, and Ajay Divakaran.
\newblock Zero-shot object detection.
\newblock In \emph{Proceedings of the European Conference on Computer Vision
  (ECCV)}, pages 384--400, 2018.

\bibitem[Carion et~al.(2020)Carion, Massa, Synnaeve, Usunier, Kirillov, and
  Zagoruyko]{DETR}
Nicolas Carion, Francisco Massa, Gabriel Synnaeve, Nicolas Usunier, Alexander
  Kirillov, and Sergey Zagoruyko.
\newblock End-to-end object detection with transformers.
\newblock In Andrea Vedaldi, Horst Bischof, Thomas Brox, and Jan{-}Michael
  Frahm, editors, \emph{Computer Vision - {ECCV} 2020 - 16th European
  Conference, Glasgow, UK, August 23-28, 2020, Proceedings, Part {I}}, volume
  12346 of \emph{Lecture Notes in Computer Science}, pages 213--229. Springer,
  2020.
\newblock \doi{10.1007/978-3-030-58452-8\_13}.
\newblock URL \url{https://doi.org/10.1007/978-3-030-58452-8\_13}.

\bibitem[Carreira and Zisserman(2017)]{I3D}
Joao Carreira and Andrew Zisserman.
\newblock Quo vadis, action recognition? a new model and the kinetics dataset.
\newblock In \emph{Proceedings of the IEEE Conference on Computer Vision and
  Pattern Recognition (CVPR)}, July 2017.

\bibitem[Chao et~al.(2018)Chao, Vijayanarasimhan, Seybold, Ross, Deng, and
  Sukthankar]{TalNet}
Yu-Wei Chao, Sudheendra Vijayanarasimhan, Bryan Seybold, David~A. Ross, Jia
  Deng, and Rahul Sukthankar.
\newblock Rethinking the faster {R}-{CNN} architecture for temporal action
  localization.
\newblock In \emph{Proceedings of the IEEE Conference on Computer Vision and
  Pattern Recognition}, 2018.

\bibitem[Deng et~al.(2009)Deng, Dong, Socher, Li, Li, and Fei-Fei]{Imagenet}
Jia Deng, Wei Dong, Richard Socher, Li-Jia Li, Kai Li, and Li~Fei-Fei.
\newblock Imagenet: A large-scale hierarchical image database.
\newblock In \emph{2009 IEEE conference on computer vision and pattern
  recognition}, pages 248--255. Ieee, 2009.

\bibitem[Duan et~al.(2019)Duan, Bai, Xie, Qi, Huang, and Tian]{CenterNet}
Kaiwen Duan, Song Bai, Lingxi Xie, Honggang Qi, Qingming Huang, and Qi~Tian.
\newblock Centernet: Keypoint triplets for object detection, 2019.
\newblock URL \url{http://arxiv.org/abs/1904.08189}.
\newblock cite arxiv:1904.08189Comment: 10 pages (including 2 pages of
  References), 7 figures, 5 tables.

\bibitem[Fabian Caba~Heilbron and Niebles(2015)]{ActivityNet}
Bernard~Ghanem Fabian Caba~Heilbron, Victor~Escorcia and Juan~Carlos Niebles.
\newblock Activitynet: A large-scale video benchmark for human activity
  understanding.
\newblock In \emph{Proceedings of the IEEE Conference on Computer Vision and
  Pattern Recognition}, pages 961--970, 2015.

\bibitem[Gemmeke et~al.(2017)Gemmeke, Ellis, Freedman, Jansen, Lawrence, Moore,
  Plakal, and Ritter]{AudioSet}
Jort~F. Gemmeke, Daniel P.~W. Ellis, Dylan Freedman, Aren Jansen, Wade
  Lawrence, R.~Channing Moore, Manoj Plakal, and Marvin Ritter.
\newblock Audio set: An ontology and human-labeled dataset for audio events.
\newblock In \emph{2017 IEEE International Conference on Acoustics, Speech and
  Signal Processing (ICASSP)}, pages 776--780, 2017.
\newblock \doi{10.1109/ICASSP.2017.7952261}.

\bibitem[Gu et~al.(2021)Gu, Lin, Kuo, and Cui]{VILD}
Xiuye Gu, Tsung{-}Yi Lin, Weicheng Kuo, and Yin Cui.
\newblock Zero-shot detection via vision and language knowledge distillation.
\newblock \emph{CoRR}, abs/2104.13921, 2021.
\newblock URL \url{https://arxiv.org/abs/2104.13921}.

\bibitem[Hershey et~al.(2017)Hershey, Chaudhuri, Ellis, Gemmeke, Jansen, Moore,
  Plakal, Platt, Saurous, Seybold, Slaney, Weiss, and Wilson]{Vggish}
Shawn Hershey, Sourish Chaudhuri, Daniel P.~W. Ellis, Jort~F. Gemmeke, Aren
  Jansen, Channing Moore, Manoj Plakal, Devin Platt, Rif~A. Saurous, Bryan
  Seybold, Malcolm Slaney, Ron Weiss, and Kevin Wilson.
\newblock Cnn architectures for large-scale audio classification.
\newblock In \emph{International Conference on Acoustics, Speech and Signal
  Processing (ICASSP)}, 2017.
\newblock URL \url{https://arxiv.org/abs/1609.09430}.

\bibitem[Huang et~al.(2017)Huang, Rathod, Sun, Zhu, Korattikara, Fathi,
  Fischer, Wojna, Song, Guadarrama, et~al.]{ODapi}
Jonathan Huang, Vivek Rathod, Chen Sun, Menglong Zhu, Anoop Korattikara,
  Alireza Fathi, Ian Fischer, Zbigniew Wojna, Yang Song, Sergio Guadarrama,
  et~al.
\newblock Speed/accuracy trade-offs for modern convolutional object detectors.
\newblock In \emph{Proceedings of the IEEE conference on computer vision and
  pattern recognition}, pages 7310--7311, 2017.

\bibitem[Idrees et~al.(2016)Idrees, Zamir, Jiang, Gorban, Laptev, Sukthankar,
  and Shah]{Thumos}
Haroon Idrees, Amir~Roshan Zamir, Yu{-}Gang Jiang, Alex Gorban, Ivan Laptev,
  Rahul Sukthankar, and Mubarak Shah.
\newblock The {THUMOS} challenge on action recognition for videos "in the
  wild".
\newblock \emph{CoRR}, abs/1604.06182, 2016.
\newblock URL \url{http://arxiv.org/abs/1604.06182}.

\bibitem[Islam and Radke(2020)]{islam2020weakly}
Ashraful Islam and Richard Radke.
\newblock Weakly supervised temporal action localization using deep metric
  learning.
\newblock In \emph{Proceedings of the IEEE/CVF Winter Conference on
  Applications of Computer Vision}, pages 547--556, 2020.

\bibitem[Jia et~al.(2021)Jia, Yang, Xia, Chen, Parekh, Pham, Le, Sung, Li, and
  Duerig]{Align}
Chao Jia, Yinfei Yang, Ye~Xia, Yi-Ting Chen, Zarana Parekh, Hieu Pham, Quoc Le,
  Yun-Hsuan Sung, Zhen Li, and Tom Duerig.
\newblock Scaling up visual and vision-language representation learning with
  noisy text supervision.
\newblock In \emph{International Conference on Machine Learning}, pages
  4904--4916. PMLR, 2021.

\bibitem[Ju et~al.(2022)Ju, Han, Zheng, Zhang, and Xie]{EfficientPrompt}
Chen Ju, Tengda Han, Kunhao Zheng, Ya~Zhang, and Weidi Xie.
\newblock Prompting visual-language models for efficient video understanding.
\newblock In \emph{European Conference on Computer Vision (ECCV)}. Springer,
  2022.

\bibitem[Kay et~al.(2017)Kay, Carreira, Simonyan, Zhang, Hillier,
  Vijayanarasimhan, Viola, Green, Back, Natsev, Suleyman, and
  Zisserman]{Kinetics}
Will Kay, Jo{\~{a}}o Carreira, Karen Simonyan, Brian Zhang, Chloe Hillier,
  Sudheendra Vijayanarasimhan, Fabio Viola, Tim Green, Trevor Back, Paul
  Natsev, Mustafa Suleyman, and Andrew Zisserman.
\newblock The kinetics human action video dataset.
\newblock \emph{CoRR}, abs/1705.06950, 2017.
\newblock URL \url{http://arxiv.org/abs/1705.06950}.

\bibitem[Li et~al.(2019)Li, Yao, Zhang, Wang, Kanhere, and Zhang]{zeroshotod4}
Zhihui Li, Lina Yao, Xiaoqin Zhang, Xianzhi Wang, Salil~S. Kanhere, and
  Huaxiang Zhang.
\newblock Zero-shot object detection with textual descriptions.
\newblock In \emph{The Thirty-Third {AAAI} Conference on Artificial
  Intelligence, {AAAI} 2019, The Thirty-First Innovative Applications of
  Artificial Intelligence Conference, {IAAI} 2019, The Ninth {AAAI} Symposium
  on Educational Advances in Artificial Intelligence, {EAAI} 2019, Honolulu,
  Hawaii, USA, January 27 - February 1, 2019}, pages 8690--8697. {AAAI} Press,
  2019.
\newblock \doi{10.1609/aaai.v33i01.33018690}.
\newblock URL \url{https://doi.org/10.1609/aaai.v33i01.33018690}.

\bibitem[Lin et~al.(2021)Lin, Xu, Luo, Wang, Tai, Wang, Li, Huang, and
  Fu]{lin2021learningafsd}
Chuming Lin, Chengming Xu, Donghao Luo, Yabiao Wang, Ying Tai, Chengjie Wang,
  Jilin Li, Feiyue Huang, and Yanwei Fu.
\newblock Learning salient boundary feature for anchor-free temporal action
  localization.
\newblock In \emph{Proceedings of the IEEE/CVF Conference on Computer Vision
  and Pattern Recognition}, pages 3320--3329, 2021.

\bibitem[Lin et~al.(2019)Lin, Liu, Li, Ding, and Wen]{BMN}
Tianwei Lin, Xiao Liu, Xin Li, Errui Ding, and Shilei Wen.
\newblock {BMN:} boundary-matching network for temporal action proposal
  generation.
\newblock In \emph{2019 {IEEE/CVF} International Conference on Computer Vision,
  {ICCV} 2019, Seoul, Korea (South), October 27 - November 2, 2019}, pages
  3888--3897. {IEEE}, 2019.
\newblock \doi{10.1109/ICCV.2019.00399}.
\newblock URL \url{https://doi.org/10.1109/ICCV.2019.00399}.

\bibitem[Liu et~al.(2021)Liu, Wang, Hu, Tang, Bai, and Bai]{Tadtr}
Xiaolong Liu, Qimeng Wang, Yao Hu, Xu~Tang, Song Bai, and Xiang Bai.
\newblock End-to-end temporal action detection with transformer, 2021.

\bibitem[Luo et~al.(2021)Luo, Ji, Zhong, Chen, Lei, Duan, and Li]{Clip4clip}
Huaishao Luo, Lei Ji, Ming Zhong, Yang Chen, Wen Lei, Nan Duan, and Tianrui Li.
\newblock Clip4clip: An empirical study of {CLIP} for end to end video clip
  retrieval.
\newblock \emph{CoRR}, abs/2104.08860, 2021.
\newblock URL \url{https://arxiv.org/abs/2104.08860}.

\bibitem[Nagrani et~al.(2021{\natexlab{a}})Nagrani, Yang, Arnab, Jansen,
  Schmid, and Sun]{MBT}
Arsha Nagrani, Shan Yang, Anurag Arnab, Aren Jansen, Cordelia Schmid, and Chen
  Sun, editors.
\newblock \emph{Attention Bottlenecks for Multimodal Fusion},
  2021{\natexlab{a}}.

\bibitem[Nagrani et~al.(2021{\natexlab{b}})Nagrani, Yang, Arnab, Jansen,
  Schmid, and Sun]{nagrani2021attention}
Arsha Nagrani, Shan Yang, Anurag Arnab, Aren Jansen, Cordelia Schmid, and Chen
  Sun.
\newblock Attention bottlenecks for multimodal fusion.
\newblock \emph{Advances in Neural Information Processing Systems}, 34,
  2021{\natexlab{b}}.

\bibitem[Nagrani et~al.(2022)Nagrani, Seo, Seybold, Hauth, Manen, Sun, and
  Schmid]{nagrani2022learningvideocc}
Arsha Nagrani, Paul~Hongsuck Seo, Bryan Seybold, Anja Hauth, Santiago Manen,
  Chen Sun, and Cordelia Schmid.
\newblock Learning audio-video modalities from image captions.
\newblock \emph{arXiv preprint arXiv:2204.00679}, 2022.

\bibitem[Narayan et~al.(2021)Narayan, Cholakkal, Hayat, Khan, Yang, and
  Shao]{narayan2021d2net}
Sanath Narayan, Hisham Cholakkal, Munawar Hayat, Fahad~Shahbaz Khan, Ming-Hsuan
  Yang, and Ling Shao.
\newblock D2-net: Weakly-supervised action localization via discriminative
  embeddings and denoised activations.
\newblock In \emph{Proceedings of the IEEE/CVF International Conference on
  Computer Vision}, pages 13608--13617, 2021.

\bibitem[Newell et~al.(2016)Newell, Yang, and Deng]{newell2016stacked}
Alejandro Newell, Kaiyu Yang, and Jia Deng.
\newblock Stacked hourglass networks for human pose estimation.
\newblock In \emph{European conference on computer vision}, pages 483--499.
  Springer, 2016.

\bibitem[Pham et~al.(2021)Pham, Dai, Ghiasi, Liu, Yu, Luong, Tan, and
  Le]{Basic}
Hieu Pham, Zihang Dai, Golnaz Ghiasi, Hanxiao Liu, Adams~Wei Yu, Minh{-}Thang
  Luong, Mingxing Tan, and Quoc~V. Le.
\newblock Combined scaling for zero-shot transfer learning.
\newblock \emph{CoRR}, abs/2111.10050, 2021.
\newblock URL \url{https://arxiv.org/abs/2111.10050}.

\bibitem[Radford et~al.(2021)Radford, Kim, Hallacy, Ramesh, Goh, Agarwal,
  Sastry, Askell, Mishkin, Clark, Krueger, and Sutskever]{CLIP}
Alec Radford, Jong~Wook Kim, Chris Hallacy, Aditya Ramesh, Gabriel Goh,
  Sandhini Agarwal, Girish Sastry, Amanda Askell, Pamela Mishkin, Jack Clark,
  Gretchen Krueger, and Ilya Sutskever.
\newblock Learning transferable visual models from natural language
  supervision.
\newblock In Marina Meila and Tong Zhang, editors, \emph{Proceedings of the
  38th International Conference on Machine Learning, {ICML} 2021, 18-24 July
  2021, Virtual Event}, volume 139 of \emph{Proceedings of Machine Learning
  Research}, pages 8748--8763. {PMLR}, 2021.
\newblock URL \url{http://proceedings.mlr.press/v139/radford21a.html}.

\bibitem[Rahman et~al.(2020)Rahman, Khan, and Barnes]{zeroshotod2}
Shafin Rahman, Salman Khan, and Nick Barnes.
\newblock Improved visual-semantic alignment for zero-shot object detection.
\newblock \emph{34th AAAI Conference on Artificial Intelligence}, 2020.

\bibitem[Ronneberger et~al.(2015)Ronneberger, Fischer, and
  Brox]{ronneberger2015u}
Olaf Ronneberger, Philipp Fischer, and Thomas Brox.
\newblock U-net: Convolutional networks for biomedical image segmentation.
\newblock In \emph{International Conference on Medical image computing and
  computer-assisted intervention}, pages 234--241. Springer, 2015.

\bibitem[Shou et~al.(2017)Shou, Chan, Zareian, Miyazawa, and Chang]{CDC}
Zheng Shou, Jonathan Chan, Alireza Zareian, Kazuyuki Miyazawa, and Shih-Fu
  Chang.
\newblock Cdc: Convolutional-de-convolutional networks for precise temporal
  action localization in untrimmed videos.
\newblock In \emph{CVPR}, 2017.

\bibitem[Stroud et~al.(2020)Stroud, Ross, Sun, Deng, Sukthankar, and
  Schmid]{WTS}
Jonathan~C. Stroud, David~A. Ross, Chen Sun, Jia Deng, Rahul Sukthankar, and
  Cordelia Schmid.
\newblock Learning video representations from textual web supervision.
\newblock \emph{CoRR}, abs/2007.14937, 2020.
\newblock URL \url{https://arxiv.org/abs/2007.14937}.

\bibitem[Su et~al.(2020)Su, Xu, Sheng, and Ouyang]{su2020pcg}
Rui Su, Dong Xu, Lu~Sheng, and Wanli Ouyang.
\newblock Pcg-tal: Progressive cross-granularity cooperation for temporal
  action localization.
\newblock \emph{IEEE Transactions on Image Processing}, 30:\penalty0
  2103--2113, 2020.

\bibitem[Sun et~al.(2017)Sun, Shrivastava, Singh, and Gupta]{JFT}
Chen Sun, Abhinav Shrivastava, Saurabh Singh, and Abhinav Gupta.
\newblock {Revisiting Unreasonable Effectiveness of Data in Deep Learning Era}.
\newblock In \emph{{IEEE International Conference on Computer Vision (ICCV)}},
  2017.

\bibitem[Tensorflow(2020)]{Yamnet}
Tensorflow.
\newblock Models/research/audioset/yamnet at master · tensorflow/models, 2020.
\newblock URL
  \url{https://github.com/tensorflow/models/tree/master/research/audioset/yamnet}.

\bibitem[Wang et~al.(2017)Wang, Xiong, Lin, and Van~Gool]{UntrimmedNets}
Limin Wang, Yuanjun Xiong, Dahua Lin, and Luc Van~Gool.
\newblock Untrimmednets for weakly supervised action recognition and detection.
\newblock In \emph{Proceedings of the IEEE conference on Computer Vision and
  Pattern Recognition}, pages 4325--4334, 2017.

\bibitem[Wang et~al.(2021)Wang, Xing, and Liu]{ActionClip}
Mengmeng Wang, Jiazheng Xing, and Yong Liu.
\newblock Actionclip: {A} new paradigm for video action recognition.
\newblock \emph{CoRR}, abs/2109.08472, 2021.
\newblock URL \url{https://arxiv.org/abs/2109.08472}.

\bibitem[Wang et~al.(2022)Wang, Zhang, Zheng, and Pan]{rcl}
Qiang Wang, Yanhao Zhang, Yun Zheng, and Pan Pan.
\newblock Rcl: Recurrent continuous localization for temporal action detection.
\newblock In \emph{2022 IEEE/CVF Conference on Computer Vision and Pattern
  Recognition (CVPR)}, pages 13556--13565, 2022.
\newblock \doi{10.1109/CVPR52688.2022.01320}.

\bibitem[Wentao~Bao(2022)]{opental}
Yu~Kong Wentao~Bao, Qi~Yu.
\newblock Opental: Towards open set temporal action localization.
\newblock In \emph{Proceedings of the IEEE/CVF Conference on Computer Vision
  and Pattern Recognition (CVPR)}, June 2022.

\bibitem[Xiong et~al.(2016)Xiong, Wang, Wang, Zhang, Song, Li, Lin, Qiao, Gool,
  and Tang]{ANet2016Classifier}
Yuanjun Xiong, Limin Wang, Zhe Wang, Bowen Zhang, Hang Song, Wei Li, Dahua Lin,
  Yu~Qiao, Luc~Van Gool, and Xiaoou Tang.
\newblock {CUHK} {\&} {ETHZ} {\&} {SIAT} submission to activitynet challenge
  2016.
\newblock \emph{CoRR}, abs/1608.00797, 2016.
\newblock URL \url{http://arxiv.org/abs/1608.00797}.

\bibitem[Xu et~al.(2017)Xu, Das, and Saenko]{RC3D}
Huijuan Xu, Abir Das, and Kate Saenko.
\newblock R-c3d: region convolutional 3d network for temporal activity
  detection.
\newblock In \emph{ICCV}, pages 5794--5803, 2017.

\bibitem[Xu et~al.(2020)Xu, Zhao, Rojas, Thabet, and Ghanem]{GTAD}
Mengmeng Xu, Chen Zhao, David~S. Rojas, Ali~K. Thabet, and Bernard Ghanem.
\newblock {G-TAD:} sub-graph localization for temporal action detection.
\newblock In \emph{2020 {IEEE/CVF} Conference on Computer Vision and Pattern
  Recognition, {CVPR} 2020, Seattle, WA, USA, June 13-19, 2020}, pages
  10153--10162. Computer Vision Foundation / {IEEE}, 2020.
\newblock \doi{10.1109/CVPR42600.2020.01017}.
\newblock URL
  \url{https://openaccess.thecvf.com/content\_CVPR\_2020/html/Xu\_G-TAD\_Sub-Graph\_Localization\_for\_Temporal\_Action\_Detection\_CVPR\_2020\_paper.html}.

\bibitem[Yang et~al.(2020)Yang, Peng, Zhang, Fu, and Han]{A2Net}
Le~Yang, Houwen Peng, Dingwen Zhang, Jianlong Fu, and Junwei Han.
\newblock Revisiting anchor mechanisms for temporal action localization.
\newblock \emph{IEEE Transactions on Image Processing}, 29:\penalty0
  8535--8548, 2020.

\bibitem[Zareian et~al.(2021)Zareian, Rosa, Hu, and Chang]{OVRCNN}
Alireza Zareian, Kevin~Dela Rosa, Derek~Hao Hu, and Shih-Fu Chang.
\newblock Open-vocabulary object detection using captions.
\newblock In \emph{Proceedings of the IEEE/CVF Conference on Computer Vision
  and Pattern Recognition}, pages 14393--14402, 2021.

\bibitem[Zeng et~al.(2019)Zeng, Huang, Tan, Rong, Zhao, Huang, and Gan]{PGCN}
Runhao Zeng, Wenbing Huang, Mingkui Tan, Yu~Rong, Peilin Zhao, Junzhou Huang,
  and Chuang Gan.
\newblock Graph convolutional networks for temporal action localization.
\newblock In \emph{The IEEE International Conference on Computer Vision
  (ICCV)}, October 2019.

\bibitem[Zhang et~al.(2022)Zhang, Wu, and Li]{ActionFormer}
Chenlin Zhang, Jianxin Wu, and Yin Li.
\newblock Actionformer: Localizing moments of actions with transformers, 2022.

\bibitem[Zhao et~al.(2020{\natexlab{a}})Zhao, Xie, Ju, Zhang, Wang, and
  Tian]{BtmUpMReg}
Peisen Zhao, Lingxi Xie, Chen Ju, Ya~Zhang, Yanfeng Wang, and Qi~Tian.
\newblock Bottom-up temporal action localization with mutual regularization.
\newblock In \emph{ECCV}, 2020{\natexlab{a}}.

\bibitem[Zhao et~al.(2020{\natexlab{b}})Zhao, Xiong, Wang, Wu, Tang, and
  Lin]{SSN}
Yue Zhao, Yuanjun Xiong, Limin Wang, Zhirong Wu, Xiaoou Tang, and Dahua Lin.
\newblock Temporal action detection with structured segment networks.
\newblock \emph{Int. J. Comput. Vis.}, 128\penalty0 (1):\penalty0 74--95,
  2020{\natexlab{b}}.
\newblock \doi{10.1007/s11263-019-01211-2}.
\newblock URL \url{https://doi.org/10.1007/s11263-019-01211-2}.

\bibitem[Zhu et~al.(2020)Zhu, Wang, and Saligrama]{zeroshotod3}
Pengkai Zhu, Hanxiao Wang, and Venkatesh Saligrama.
\newblock Don't even look once: Synthesizing features for zero-shot detection.
\newblock In \emph{2020 {IEEE/CVF} Conference on Computer Vision and Pattern
  Recognition, {CVPR} 2020, Seattle, WA, USA, June 13-19, 2020}, pages
  11690--11699, 2020.
\newblock \doi{10.1109/CVPR42600.2020.01171}.
\newblock URL
  \url{https://openaccess.thecvf.com/content\_CVPR\_2020/html/Zhu\_Dont\_Even\_Look\_Once\_Synthesizing\_Features\_for\_Zero-Shot\_Detection\_CVPR\_2020\_paper.html}.

\end{thebibliography}

\clearpage

\appendix


\section{Class-Agnostic Detection Architectures}

\begin{figure}
    \centering
    \includegraphics[width=0.8\textwidth]{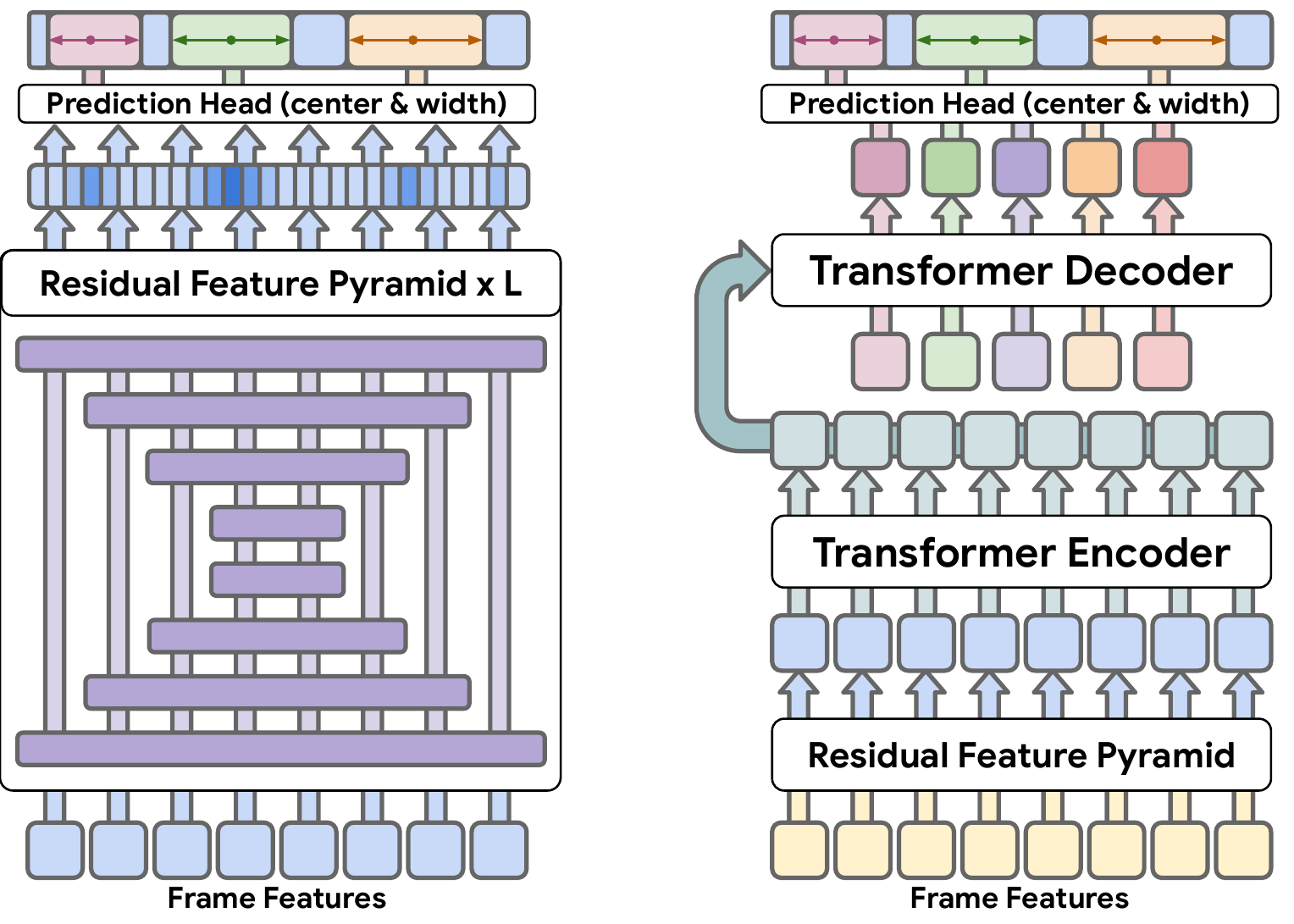}
    \caption{
    Class-Agnostic detection architectures used in our experiments.
    (Left) Illustration of the CenterNet\cite{CenterNet} based class-agnostic detector, 
    which applies a residual feature pyramid (RFP) followed by action prediction heads.
    (Right) Illustration of the DETR\cite{DETR} based model, which first applies RFP to input features
    followed by transformer encoder and decoder layers. See text for details. 
    }
    \label{fig:det_arch}
\end{figure}

Figure~\ref{fig:det_arch} is an overview of our class-agnostic detection architectures that we describe in detail below.

\subsection{Residual Feature Pyramid}
We construct a residual feature pyramid inspired by \cite{newell2016stacked} to mix temporal information and generate appropriate inputs for detection heads. We construct and stack multiple U-Net\cite{ronneberger2015u}-style architectures on top of one another. We use residual connections extensively throughout the network and across stacked towers at the same resolution. We call our particular configuration a Residual Feature Pyramid (RFP).

\begin{figure}
    \centering
    \includegraphics[width=\textwidth]{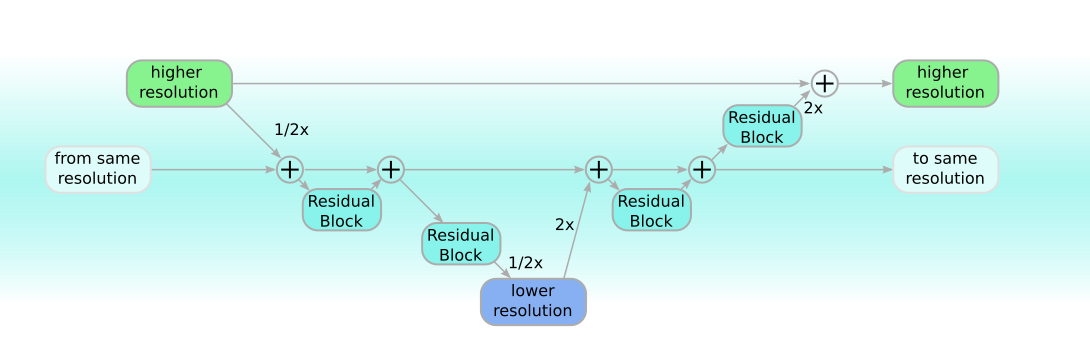}
    \caption{Diagram of one resolution in a Residual Feature Pyramid. The shaded, middle portion of diagram is at a single resolution. The top is a higher resolution, and the bottom is lower resolution. Information travels from left to right and this covers both the descending in resolution and ascending portions. Each $\bigoplus$ represents a residual addition.}
    \label{fig:rfp}
\end{figure}

Supplemental Figure~\ref{fig:rfp} details one resolution of the RFP in the descending then ascending directions. Inputs for this resolution come in from downsampling the higher resolution and features at the same resolution from previous RFPs in the stack. The input features are summed and passed to two residual blocks. After two blocks the features are downsampled via a strided convolution and passed to a similar, nested processing stage at a lower resolution. The upsampled outputs of the lower resolution are summed with the previous residual values and passed to two more residual blocks. The final output is then upsampled via strided transposed convolution and processed by the higher resolution. Our residual blocks are a 0) Input storage, 1) ReLU activation, 2) Convolution, 3) LayerNormalization, 4) ReLU activation, 5) Convolution, 6) LayerNormalization, and 7) Residual addition to the stored input. If downsampling or upsampling, (5) is a stride 2 convolution or transposed convolution and the residual inputs (7) are processed by an additional strided convolution to match the resolution and number of dimensions.

The hyperparameters for our stacked RFP are as follows. We stack 4 descending and ascending RFPs on top of each other. Each RFP has 5 resolutions. At the highest, full resolution each convolution has 512 units. All lower resolutions use 128 units. Each kernel has a window size of 5.

\subsection{DETR}
We adapt the DETR\cite{DETR} architecture proposed for object detection to temporal action detection (TAD) by treating TAD as a 1-d version of object detection. The DETR model consists of a transformer encoder and a decoder and treats object detection as a set prediction problem. The model takes as input a fixed number of learned object query vectors and uses prediction heads on top of the transformer outputs to predict bounding boxes and object classification scores directly. 

In our adaptation, we predict segment start and end indices and action classification scores from the transformer decoder's output with the following modifications. (1) We use a residual feature pyramid architecture to extract features, (2) we perform SOI (Segment of Interest) pooling as defined in\cite{TalNet} using the predicted segments and concatenate the pooled features with the transformer output before classifying the action, (3) we use learned positional encodings and perform hungarian matching on a combination of the L1 distance and IOU scores as done in \cite{Tadtr}. 

\textbf{Hyperparameters.} We use $2$ hidden layers for both the encoder and decoder, hidden size of $1024$ and $16$ attention heads. The prediction head consists of $2$ linear layers of hidden size $1024$. We set the number of segment queries to $64$ and perform NMS to obtain the final set of segment detections.  

\subsection{CenterNet}
We adapt the CenterNet\cite{CenterNet} architecture proposed for object detection to temporalaction detection (TAD) by treating TAD as a 1-d version of object detection. CenterNet produces three outputs at every input point: a logit distribution for each class label, a width if any segment starts at that point, and an offset that shifts the center location slightly if needed. Each prediction is produced by a 2-layer convolutional network with $256$ units on top of an RFP feature extractor. As in \cite{CenterNet}, we predict segments at the full resolution between stacked RFP and apply auxiliary losses at each resolution during training. During inference we only use segments from the last RFP stack. The loss for the width and offset are absolute error. The loss for the classification logits is the penalty reduced focal loss described in \cite{CenterNet}. We weight the width by $0.1$ relative to the other losses.

\section{Additional Results}

\subsection{Class-Agnostic Detection}

\subsubsection{Our random splits are statistically similar to \cite{EfficientPrompt}}.

\cite{EfficientPrompt} uses 10 random splits, and we use a different set of 12 random splits for 75/25 and 4 for 50/50. To show that our splits are statistically similar those in \cite{EfficientPrompt}, we compare the means and standard errors on class-agnostic detection on ActivityNet with both sets of 75/25 splits in Table~\ref{tab:anet_ep_proposals_supp}. As expected, we find that the means are often within one standard error of each other for nearly every possible comparison.

\begin{table}[t]
    \centering
    \scalebox{0.7}{
    \begin{tabular}{|c|c|c|c|c|c|c|c|}
    \hline
    Model & mAP@0.5 & mAP@0.75 & mAP@0.95 & mAP@avg & AR@10 & AR@50 & AR@100 \\
    \hline

Clip b/32 - our splits & 48.4 $\pm$ 0.6 & 25.7 $\pm$ 0.7 & 3.3 $\pm$ 0.3 & 27.2 $\pm$ 0.5 & 41.2 $\pm$ 0.5 & 47.9 $\pm$ 0.4 & 47.9 $\pm$ 0.4\\
Clip b/32 - \cite{EfficientPrompt} splits & 48.7 $\pm$ 1.0 & 25.7 $\pm$ 0.7 & 3.4 $\pm$ 0.3 & 27.3 $\pm$ 0.7 & 41.2 $\pm$ 0.6 & 47.2 $\pm$ 0.6 & 47.2 $\pm$ 0.6\\
\hline
Clip b/16 - our splits & 49.0 $\pm$ 0.8 & 26.1 $\pm$ 0.6 & 3.2 $\pm$ 0.3 & 27.6 $\pm$ 0.5 & 42.2 $\pm$ 0.6 & 49.1 $\pm$ 0.7 & 49.1 $\pm$ 0.7\\
Clip b/16 - \cite{EfficientPrompt} splits & 49.1 $\pm$ 1.0 & 26.2 $\pm$ 0.8 & 3.5 $\pm$ 0.3 & 27.7 $\pm$ 0.7 & 41.6 $\pm$ 0.5 & 48.4 $\pm$ 0.5 & 48.4 $\pm$ 0.5\\
\hline
Clip l/14 - our splits & 49.6 $\pm$ 0.6 & 26.7 $\pm$ 0.5 & 3.8 $\pm$ 0.3 & 28.3 $\pm$ 0.4 & 42.4 $\pm$ 0.4 & 48.3 $\pm$ 0.5 & 48.3 $\pm$ 0.5\\
Clip l/14 - \cite{EfficientPrompt} splits & 50.2 $\pm$ 1.0 & 26.7 $\pm$ 0.8 & 3.6 $\pm$ 0.3 & 28.4 $\pm$ 0.8 & 42.2 $\pm$ 0.6 & 48.3 $\pm$ 0.7 & 48.3 $\pm$ 0.7\\
\hline
Align - our splits & 49.1 $\pm$ 0.6 & 26.4 $\pm$ 0.5 & 3.4 $\pm$ 0.3 & 27.8 $\pm$ 0.5 & 42.2 $\pm$ 0.4 & 47.8 $\pm$ 0.5 & 47.8 $\pm$ 0.5\\
Align - \cite{EfficientPrompt} splits & 49.8 $\pm$ 1.0 & 26.3 $\pm$ 0.8 & 3.8 $\pm$ 0.4 & 28.0 $\pm$ 0.8 & 42.0 $\pm$ 0.6 & 47.4 $\pm$ 0.5 & 47.4 $\pm$ 0.5\\
\hline
Basic - our splits & 50.3 $\pm$ 0.7 & 27.5 $\pm$ 0.6 & 3.3 $\pm$ 0.3 & 28.8 $\pm$ 0.5 & 43.3 $\pm$ 0.5 & 49.6 $\pm$ 0.3 & 49.6 $\pm$ 0.3\\
Basic - \cite{EfficientPrompt} splits & 51.6 $\pm$ 0.9 & 27.8 $\pm$ 0.8 & 3.4 $\pm$ 0.2 & 29.3 $\pm$ 0.7 & 43.5 $\pm$ 0.6 & 48.9 $\pm$ 0.5 & 48.9 $\pm$ 0.5\\
\hline

    \end{tabular}
    }
    \caption{
ActivityNet Class-Agnostic Detection Results comparing our splits to \cite{EfficientPrompt}.
 Performance of class-agnostic detectors with different input features. mAP@0.5IoUs is the mean Average 
Precision at 0.5 Intersection over Union. mAP@avg is the average of
mAP@0.50:0.05:0.95. AR@10 is Average Recall at 10. Aggregates are means 
$\pm$ standard error across splits.}
    \label{tab:anet_ep_proposals_supp}
\end{table}

\subsubsection{Variability on Thumos}

Evaluation performance on Thumos moves in large, discrete jumps as shown in Supplementary Figure~\ref{fig:supp-variability}. This examples shows one evaluation curve for mAP@0.5 IoU, but the dominant behavior of large, discrete jumps in evaluation performance is consistent across many runs and all mAP metrics. This discrete behavior is because here are only five classes in the 75/25 splits of Thumos and when the class-agnostic model learns or unlearns segments applicable to that class, performance jumps. The large jumps are related to the standard error of the mean mAP being ten times as large on Thumos as on ActivityNet. 

\begin{figure}
\centering
\includegraphics[width=0.5\textwidth]{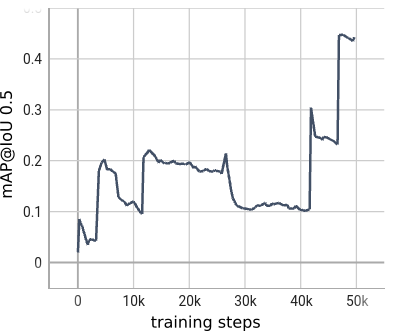}
\caption{
mAP@0.5 IoU evaluation performance on Thumos moves in large, discrete jumps.
}
\label{fig:supp-variability}
\end{figure}

\subsection{Fully supervised detection with video transformers relies on classification}

Table~\ref{tab:anet_proposals_supp} mirrors Table 2 in the main text but only shows ActivityNet performance and adds fully-supervised detection metrics. The most interesting comparisons are between the fully-supervised mAPs and the class-agnostic mAPs. For ITCE features, both metrics increase with model size and class-agnostic metrics are typically slightly better than fully-supervised ones. In contrast, for traditional features (I3D and audio models), the fully-supervised performance is much worse than class-agnostic performance. The opposite trend is true where recent vision transformer models~\cite{MBT,Vivit} perform well for fully-supervised detection, but relatively poorly for class-agnostic detection.

\begin{table}[t]
    \centering
    \scalebox{0.6}{
    \begin{tabular}{|c|c|c|c|c|c|c|c|c|c|c|}
    \hline
  \multirow{3}{*}{Model} & \multicolumn{2}{c|}{Supervised} &  \multicolumn{2}{c|}{75/25 splits}   & \multicolumn{2}{c|}{Intermediate split} & \multicolumn{2}{c|}{Smart split}  & \multicolumn{2}{c|}{50/50 splits} \\
 & \multicolumn{2}{c|}{(200 class)}  &  \multicolumn{2}{c|}{average (n=12)}   & \multicolumn{2}{c|}{}  & \multicolumn{2}{c|}{}  & \multicolumn{2}{c|}{average (n=4)} \\
 \cline{2-11}
 & mAP@avg & AR@10 & AP@avg & AR@10 & AP@avg & AR@10 & AP@avg & AR@10 & AP@avg & AR@10\\
\hline
CLIP b/32 & 21.5 & 43.2 & 27.2 $\pm$ 1.0 & 41.3 $\pm$ 0.9 & 28.4 & 42.1 & 29.2 & 43.0 & 25.6 $\pm$ 0.4 & 38.6 $\pm$ 0.9\\
CLIP b/16 & 24.0 & 44.5 & 27.6 $\pm$ 1.0 & 42.4 $\pm$ 1.2 & 27.9 & 41.2 & 30.7 & 44.0 & 25.8 $\pm$ 0.7 & 40.0 $\pm$ 1.3\\
CLIP l/14 & 26.1 & 44.3 & 28.2 $\pm$ 0.8 & 42.5 $\pm$ 0.9 & 28.7 & 43.2 & 31.3 & 43.4 & 26.6 $\pm$ 0.6 & 40.1 $\pm$ 0.7\\
ALIGN & 25.9 & 43.3 & 27.8 $\pm$ 0.9 & 42.3 $\pm$ 0.8 & 28.7 & 41.9 & 30.3 & 43.0 & 26.1 $\pm$ 0.9 & 40.1 $\pm$ 1.0\\
BASIC & \textbf{29.4} & \textbf{45.7} & \textbf{28.8 $\pm$ 0.9} & \textbf{43.3 $\pm$ 0.8} & 28.6 & 43.0 & \textbf{32.0} & \textbf{44.9} & \textbf{27.0 $\pm$ 0.8} & \textbf{41.0 $\pm$ 1.1}\\
I3D RGB+FLOW & 22.6 & 44.6 & 28.4 $\pm$ 0.9 & 42.3 $\pm$ 0.8 & \textbf{28.8} & \textbf{44.2} & 30.4 & 43.1 & 26.5 $\pm$ 0.5 & 40.4 $\pm$ 0.7\\
I3D RGB & 21.0 & 43.6 & 27.2 $\pm$ 0.7 & 41.3 $\pm$ 0.7 & 27.5 & 41.4 & 30.7 & 43.8 & 25.1 $\pm$ 0.3 & 38.9 $\pm$ 0.3\\
I3D FLOW & 17.7 & 44.6 & 28.3 $\pm$ 1.0 & 42.2 $\pm$ 1.0 & 28.8 & 40.9 & 29.6 & 42.5 & 25.6 $\pm$ 0.5 & 40.3 $\pm$ 01.3\\
YAMNET & 6.6 & 36.8 & 22.7 $\pm$ 0.6 & 36.6 $\pm$ 0.6 & 23.1 & 37.8 & 25.5 & 40.1 & 20.9 $\pm$ 1.0 & 34.9 $\pm$ 01.7\\
VGGISH & 7.3 & 37.0 & 23.0 $\pm$ 0.8 & 36.9 $\pm$ 0.9 & 23.2 & 36.7 & 25.0 & 38.8 & 21.3 $\pm$ 0.5 & 35.6 $\pm$ 1.1\\
VIVIT & 25.5 & 41.7 & 22.4 $\pm$ 0.8 & 37.1 $\pm$ 1.1 & 22.7 & 37.8 & 24.8 & 40.0 & 20.0 $\pm$ 0.4 & 34.2 $\pm$ 1.1\\
MBT-v & 25.3 & 42.4 & 23.5 $\pm$ 0.7 & 38.5 $\pm$ 0.8 & 22.7 & 37.3 & 25.6 & 40.8 & 21.8 $\pm$ 0.4 & 36.9 $\pm$ 0.9\\
MBT-a & 3.1 & 33.8 & 18.1 $\pm$ 0.8 & 34.1 $\pm$ 1.0 & 19.2 & 35.3 & 20.0 & 35.1 & 17.9 $\pm$ 0.3 & 33.6 $\pm$ 0.5\\
\hline
    \end{tabular}}
    \vspace{0.1in}
    \caption{ActivityNet Class-Agnostic Detection Results. Performance of class-agnostic detectors with different input features. mAP@avg is the average mAP over IoUs 0.5:0.05:0.95. AR@10 is Average Recall at 10. Data is grouped as columns for supervised, closed-vocabulary TAD, 75/25 random splits (n=12) (along with the Intermediate and Smart splits), and 50/50 random splits (n=4). Aggregates are means $\pm$ standard error across splits. See supplemental for additional metrics.}
    \label{tab:anet_proposals_supp}
\end{table}


\subsection{Feature Ensembles for Class Agnostic Detection on limited splits}

To determine which feature ensembles improve performance, we began by testing combinations on a subset of splits in Table~\ref{tab:anet_proposal_merge}. The two splits are the Intermediate split, which was one of the random splits that happened to give intermediate performance, and the Smart split. After testing many combinations, we took the three most synergistic features, I3D\_FLOW $+$ I3D\_RGB $+$ VGGISH and computed how much benefit they added to different ITCE features in Table~\ref{tab:merged_feature_proposals}

\begin{table}[t]
    \centering
    \scalebox{0.68}{
    \begin{tabular}{|c|c|c|c|c|c|c|c|c|c|c|c|c|}
    \hline
    \multirow{3}{*}{Feature}& \multicolumn{10}{c|}{Image-text Model} & \multicolumn{2}{c|}{Delta}\\
    \cline{2-13}
         	& \multicolumn{2}{c|}{CLIP-b/32}	&	\multicolumn{2}{c|}{CLIP-b/16}	&	\multicolumn{2}{c|}{CLIP-l/14} & \multicolumn{2}{c|}{ALIGN} & \multicolumn{2}{c|}{BASIC} & Mean & Stderr \\	
         \cline{2-11}
	& Int.	& Smart	 & Int.	 & Smart & Int. &	Smart &	Int. &	Smart	& Int. &	Smart &	 &	(n=10) \\
\hline
[image-text alone] &	28.50 &	29.70 &	29.00 &	30.40 &	28.80 &	31.50 &	28.50 &	29.30 &	29.30&	31.70&	N/A	& N/A \\
i3d\_flow&	29.24&	30.60&	28.99&	30.89&	28.95&	29.98&	29.42&	31.62&	30.42&	32.20&	0.56&	0.32\\
i3d\_rgb	&28.80&	30.74&	28.35&	30.47&	28.35&	30.83&	29.05&	30.50&	29.80&	31.17&	0.14&	0.23\\
yamnet	&28.46&	30.32&	28.14&	30.35&	29.52&	31.01&	28.39&	32.03&	29.06&	32.10&	0.27&	0.33\\
vggish&	28.40&	29.69&	29.42&	31.13&	29.83&	31.53&	29.27&	30.35&	29.85&	31.95&	0.47&	0.14\\
i3d\_flow,yamnet&	29.20&	31.01&	29.82&	31.54&	28.75&	31.66&	28.97&	30.86&	30.11&	32.97&	0.82&	0.17\\
i3d\_flow,vggish&	29.13&	31.53&	28.94&	31.16&	30.16&	31.80&	29.27&	32.31&	29.24&	32.25&	0.91&	0.31\\
i3d\_flow,i3d\_rgb&	28.63&	30.84&	29.30&	30.57&	29.93&	32.09&	29.51&	31.35&	29.54&	31.88&	0.69&	0.21\\
i3d\_flow,i3d\_rgb,yamnet&	28.63&	31.49&	29.43&	30.88&	28.63&	31.80&	29.84&	31.07&	30.03&	32.19&	0.73&	0.23\\
i3d\_flow,i3d\_rgb,vggish&	29.23&	30.91&	29.93&	31.03&	29.67&	31.34&	29.58&	31.95&	29.66&	31.84&	0.84&	0.26\\
\hline
    \end{tabular}}
    \vspace{0.1in}
    \caption{ActivityNet Class-Agnostic Temporal Action Detection on Ensembled Features. Comparisons on Intermediate (Int.) and Smart splits of ActivityNet across different image-text embeddings (columns) and different ensemble features (rows). The rightmost columns are the mean and standard error across column values relative to the [image-text alone] row measuring the added benefit of using these ensemble features. See supplemental for additional metrics.}
    \label{tab:anet_proposal_merge}
\end{table}

\begin{table}[!h]
    \centering
    \scalebox{0.9}{
    \begin{tabular}{|c|c|c|c|c|c|c|}
    \multicolumn{7}{c}{\large ActivityNet - 75/25 split average (n=12)} \\
    \hline
     Embedding & CLIP-b/32 & CLIP-b/16 & CLIP-l/14 & ALIGN & BASIC & avg. diff. \\
     \hline
B & 27.2 $\pm$ 1 & 27.6 $\pm$ 1 & 28.2 $\pm$ 0.8 & 27.8 $\pm$ 0.9 & 28.8 $\pm$ 0.9 & N/A\\
B+F & 28.2 $\pm$ 0.5 & 28.4 $\pm$ 0.4 & 28.6 $\pm$ 0.5 & 29.1 $\pm$ 0.5 & 28.9 $\pm$ 0.5 & 0.7\\
B+F+A & 28.2 $\pm$ 0.5 & 28.3 $\pm$ 0.4 & 28.8 $\pm$ 0.4 & 29.2 $\pm$ 0.5 & 28.6 $\pm$ 0.5 & 0.7\\
B+F+A+V & 28.5 $\pm$ 0.6 & 28.9 $\pm$ 0.5 & 29 $\pm$ 0.5 & 29.1 $\pm$ 0.5 & 29 $\pm$ 0.5 & 1.0\\
\hline
 \multicolumn{7}{c}{\large ActivityNet - Smart} \\
 \hline
B & 29.2 & 30.7 & 31.3 & 30.3 & 32.0 & N/A\\
B+F & 30.6 & 30.9 & 30.0 & 31.6 & 32.2 & 0.3\\
B+F+A & 31.5 & 31.2 & 31.8 & 32.3 & 32.2 & 1.1\\
B+F+A+V & 30.9 & 31.0 & 31.3 & 32.0 & 31.8 & 0.7\\
\hline
 \multicolumn{7}{c}{\large Thumos}  \\
 \hline
B & 32.2 $\pm$ 7.7 & 32.1 $\pm$ 8 & 34.8 $\pm$ 7.9 & 31.7 $\pm$ 7.9 & 33.7 $\pm$ 7.9 & N/A\\
B+F & 27.6 $\pm$ 3.8 & 33.2 $\pm$ 4.9 & 39.7 $\pm$ 3.6 & 38.4 $\pm$ 2.9 & 33.4 $\pm$ 4.2 & 1.1\\
B+F+A & 30.8 $\pm$ 4.4 & 35.8 $\pm$ 3.7 & 40.5 $\pm$ 3.9 & 30.4 $\pm$ 4.6 & 28.8 $\pm$ 4.1 & -0.1\\
B+F+A+V & 35.1 $\pm$ 4.1 & 40.8 $\pm$ 3.7 & 33.8 $\pm$ 3.3 & 38.9 $\pm$ 2.9 & 36.2 $\pm$ 3.5 & 3.6 \\
\hline
    \end{tabular}}
    \vspace{0.1in}
  \caption{Class-agnostic Detection with Feature Ensembles. The benefits of ensembling the best features from Table \ref{tab:anet_proposal_merge} for class-agnostic detection: B - the base image-text feature alone; B+F - adding I3D\_FLOW; B+F+A - adding VGGISH audio; B+F+A+V - adding I3D\_RGB video features. ActivityNet split average is mAP@0.5:0.05:0.95. Thumos split average is mAP@0.3:0.1:0.7. See supplemental for additional metrics.}
    \label{tab:merged_feature_proposals}
\end{table}

\section{Open Vocabulary Detection}

\subsection{Expanded Comparison of Open Vocabulary Detection to State of the Art.}

We present a comparison of our proposed open-vocabulary detection method to  
other state-of-the-art methods on ActivityNet-1.3 in table \ref{tab:anet_sota_comparison}. 
In addition to a comparison against the concurrent open-vocabulary method EfficientPrompt \cite{EfficientPrompt},
we also present results of recent fully supervised action detection methods, 
including methods which utilize an ensemble of action classifiers.
We observe that our method significantly outperforms in the open-vocabulary setting. Our approach is even competitive with state-of-the-art fully-supervised self-contained methods with random splits and outperforms them by a large amount on the Smart split. Because the data distribution for training and testing classes matches in the fully-supervised setting, our Smart split results are the more appropriate comparison. Our results on the Smart split are event approaching ensembles of multiple action classifiers specific to the ActivityNet data set. 
Note, that the results presented here are from our models with the best combination of input features,
and a full breakdown of results with different input feature ensembles can be found in section \ref{subsec:exhaustive_tables}.

\begin{table*}[tb]
\centering
\begin{tabular}{|l|c|c|c|c|}
\hline
Method & mAP@0.5 & mAP@0.75 & mAP@0.95 & mAP@avg \\
\hline \hline
\multicolumn{5}{|l|}{\textit{\textbf{Open-Vocabulary methods}}}   \\ \hline
EP-full \cite{EfficientPrompt} - 75/25 split & 37.6 & 22.9 & 3.8 & 23.1 \\
Ours (best) - 75/25 split                           & \textbf{46.6$\pm$0.5} & \textbf{28.4$\pm$0.6} & - & \textbf{28.8$\pm$0.5} \\
\hline
EP-full \cite{EfficientPrompt} - 50/50 split & 32.0 & 19.3 & 2.9 & 19.6 \\
Ours (best) - 50/50 split                           & \textbf{45.3$\pm$0.5} & \textbf{27.0$\pm$0.7} & - & \textbf{27.6$\pm$0.5} \\
\hline
Ours (best) - Smart split                           & \textbf{51.6} & \textbf{31.3} &   - & \textbf{32.1} \\

\hline \hline
\multicolumn{5}{|l|}{\textit{\textbf{Self-contained methods}}}   \\ \hline
CDC~\cite{CDC}        & \textbf{43.83} & 25.88 & 0.21 & 22.77 \\
R-C3D~\cite{RC3D}     & 26.80 &     - &    - &     - \\
SSN~\cite{SSN}        & 39.12 & 23.48 & 5.49 & 23.98 \\
TAL-Net~\cite{TalNet} & 38.23 & 18.30 & 1.30 & 20.22 \\
P-GCN~\cite{PGCN}     & 42.90 & 28.14 & 2.47 & 26.99 \\
TadTR \cite{Tadtr}    & 40.85 & \textbf{28.44} & \textbf{7.84} & \textbf{27.75} \\

\hline \hline
\multicolumn{5}{|l|}{\textbf{\textit{Combined with an ensemble of action classifiers}~\cite{ANet2016Classifier}}} \\ \hline
P-GCN~\cite{PGCN}           & 48.26 & 33.16 & 3.27 & 31.11 \\
MR~\cite{BtmUpMReg}         & 43.47 & 33.91 & \textbf{9.21} & 30.12 \\
A2Net~\cite{A2Net}          & 43.55 & 28.69 & 3.70 & 27.75 \\
BMN~\cite{BMN}              & 50.07 & 34.78 & 8.29 & 33.85 \\
G-TAD~\cite{GTAD}           & 50.36 & 34.60 & 9.02 & 34.09 \\
TadTR+ensemble \cite{Tadtr} & 49.08 & 32.58 & 8.49 & 32.27 \\
TadTR+BMN \cite{Tadtr}      & \textbf{50.51} & \textbf{35.35} & 8.18 & \textbf{34.55} \\
\hline
\end{tabular}
\caption{Comparison of different methods on ActivityNet-1.3.
Methods in the first group are open-vocabulary methods, 
those in the second group are self-contained methods trained in the traditional supervised setting,
and the third group are supervised methods combined with an ensemble of action classifiers~\cite{ANet2016Classifier}.}
\label{tab:anet_sota_comparison}
\end{table*}

%

\subsection{MBT Video Model}
We also evaluate a segment-level classification model for open vocabulary TAD using the recently proposed Multi-modal Bottleneck Transformer (MBT) \cite{MBT}. MBT is a segment-level model that takes $32$ video frames as input and learns segment embeddings that are close to textual descriptions of videos. The base architecture of MBT is equivalent to Clip b/32 in terms of the number of parameters. We use the WTS-70M\cite{WTS} dataset which is a weakly supervised dataset containing internet video titles to train the model. Table~\ref{tab:anet_e2e_supp} provides results using the MBT model as the open vocabulary classifier. MBT performs better than its equivalent Clip models suggesting that segment level embeddings encode more information than frame-level ones. However, it is still weaker than stronger image architectures such as Align/Basic. We plan to further explore this model with stronger architectures and additional features such as audio in the future.  

\subsection{Exhaustive Tables}
\label{subsec:exhaustive_tables}

Because we test so many combinations of features over different splits, the tables of results with all of the standard metrics were too cluttered for a readable main text. To still provide all of the standard metrics, we provide these exhaustive tables in the supplemental that add columns that were omitted in the main text and adjust the layout.

Exhaustive Table Index:

\begin{itemize}

\item Table~\ref{tab:anet_proposals_supp} - ActivityNet Class-Agnostic Detection Results. 

\item Table~\ref{tab:thumos_proposals_supp} - Thumos Class-Agnostic Detection Results. 

\item Table~\ref{tab:anet_e2e_supp} - ActivityNet   Open  Vocabulary  Detection  Results.

\item Table~\ref{tab:thumos_e2e_supp} - Thumos   Open  Vocabulary  Detection  Results.

\item Table~\ref{tab:anet_merged_subset_proposals_supp} - ActivityNet Class-Agnostic Detection Results for Feature Ensembles on a Subset of Splits.

\item Table~\ref{tab:anet_merged_proposals_supp} - ActivityNet Class-Agnostic Detection Results for Feature Ensembles.

\item Table~\ref{tab:thumos_merged_proposals_supp} - Thumos Class-Agnostic Detection Results for Feature Ensembles.

\item Table~\ref{tab:anet_e2e_merged_supp} - ActivityNet Open  Vocabulary  Detection  Results for Feature Ensembles.

\item Table~\ref{tab:thumos_e2e_merged_supp} - Thumos Open  Vocabulary  Detection  Results for Feature Ensembles.

\item Table~\ref{tab:anet_7525_splits_supp} - ActivityNet Random 75/25 Splits.

\item Table~\ref{tab:anet_5050_splits_supp} - ActivityNet Random 50/50 Splits.

\item Table~\ref{tab:thumos_7525_splits_supp} - Thumos Random 75/25 Splits.

\item Table~\ref{tab:thumos_5050_splits_supp} - Thumos Random 50/50 Splits.
\end{itemize}

\begin{table}[t]
    \centering
    \scalebox{0.6}{
\begin{tabular}{|c|c|c|c|c|c|c|c|}
    \hline
    Model & mAP@0.5 & mAP@0.75 & mAP@0.95 & mAP@avg & AR@10 & AR@50 & AR@100 \\
    \hline

Clip b/32 - 75/25 & 48.4 $\pm$ 0.6 & 25.7 $\pm$ 0.7 & 3.3 $\pm$ 0.3 & 27.2 $\pm$ 0.5 & 41.2 $\pm$ 0.5 & 47.9 $\pm$ 0.4 & 47.9 $\pm$ 0.4\\
Clip b/32 - Intermediate & 49.5 & 26.9 & 4.2 & 28.4 & 42.1 & 49.5 & 49.5\\
Clip b/32 - Smart & 50.4 & 27.4 & 4.2 & 29.2 & 42.5 & 50.0 & 50.0\\
Clip b/32 - 50/50 & 47.5 $\pm$ 0.7 & 24.5 $\pm$ 1.3 & 2.8 $\pm$ 0.7 & 26.2 $\pm$ 1.1 & 39.8 $\pm$ 0.9 & 47.2 $\pm$ 1.3 & 47.2 $\pm$ 1.3\\
\hline
Clip b/16 - 75/25 & 49.0 $\pm$ 0.8 & 26.1 $\pm$ 0.6 & 3.2 $\pm$ 0.3 & 27.6 $\pm$ 0.5 & 42.2 $\pm$ 0.6 & 49.1 $\pm$ 0.7 & 49.1 $\pm$ 0.7\\
Clip b/16 - Intermediate & 50.1 & 26.0 & 3.9 & 27.9 & 41.2 & 47.6 & 47.6\\
Clip b/16 - Smart & 52.0 & 29.1 & 5.1 & 30.7 & 43.9 & 50.4 & 50.4\\
\hline

    \end{tabular}
    }
    \caption{
ActivityNet Class-Agnostic Detection Results. Performance of class-agnostic 
detectors with different input features. mAP@0.5IoUs is the mean Average 
Precision at 0.5 Intersection over Union. mAP@avg is the average of
mAP@0.50:0.05:0.95. AR@10 is Average Recall at 10. Aggregates are means 
$\pm$ standard error across splits.}
    \label{tab:anet_proposals_supp}
\end{table}

\begin{table}[t]
    \centering
    \scalebox{0.55}{
    \begin{tabular}{|c|c|c|c|c|c|c|c|c|c|}
    \hline
    Model & mAP@0.3 & mAP@0.4 & mAP@0.5 & mAP@0.6 & mAP@0.7 & mAP@avg & AR@10 & AR@50 & AR@100 \\
    \hline

Clip b/32 - 75/25 & 49.6 $\pm$ 3.9 & 41.3 $\pm$ 4.0 & 32.2 $\pm$ 4.0 & 23.6 $\pm$ 4.2 & 16.4 $\pm$ 4.0 & 32.6 $\pm$ 3.9 & 14.3 $\pm$ 1.7 & 28.2 $\pm$ 2.8 & 32.7 $\pm$ 2.7\\
Clip b/32 - 50/50 & 33.4 $\pm$ 7.9 & 25.7 $\pm$ 7.1 & 17.8 $\pm$ 6.1 & 10.6 $\pm$ 4.4 & 5.1 $\pm$ 2.4 & 18.5 $\pm$ 5.5 & 8.8 $\pm$ 3.2 & 20.3 $\pm$ 4.1 & 27.0 $\pm$ 3.9\\
\hline
Clip b/16 - 75/25 & 46.5 $\pm$ 4.1 & 39.6 $\pm$ 4.1 & 32.1 $\pm$ 4.2 & 24.3 $\pm$ 4.1 & 17.7 $\pm$ 4.0 & 32.0 $\pm$ 4.0 & 14.1 $\pm$ 1.8 & 29.5 $\pm$ 2.5 & 34.5 $\pm$ 2.5\\
Clip b/16 - 50/50 & 41.2 $\pm$ 5.8 & 32.0 $\pm$ 5.6 & 21.9 $\pm$ 4.7 & 12.4 $\pm$ 3.4 & 5.5 $\pm$ 1.8 & 22.6 $\pm$ 4.1 & 9.7 $\pm$ 1.9 & 21.7 $\pm$ 3.2 & 25.9 $\pm$ 2.9\\
\hline
Clip l/14 - 75/25 & 53.4 $\pm$ 3.5 & 44.4 $\pm$ 3.9 & 34.8 $\pm$ 4.1 & 25.8 $\pm$ 4.2 & 17.6 $\pm$ 4.0 & 35.2 $\pm$ 3.8 & 15.6 $\pm$ 1.7 & 28.8 $\pm$ 2.8 & 33.4 $\pm$ 2.8\\
Clip l/14 - 50/50 & 47.9 $\pm$ 4.6 & 37.0 $\pm$ 5.8 & 25.0 $\pm$ 5.7 & 14.8 $\pm$ 4.9 & 7.9 $\pm$ 3.7 & 26.5 $\pm$ 4.7 & 10.5 $\pm$ 2.0 & 21.8 $\pm$ 4.2 & 25.9 $\pm$ 4.3\\
\hline
Align - 75/25 & 47.3 $\pm$ 4.2 & 39.8 $\pm$ 4.2 & 31.7 $\pm$ 4.1 & 23.0 $\pm$ 4.0 & 15.8 $\pm$ 3.9 & 31.5 $\pm$ 3.9 & 14.2 $\pm$ 1.7 & 29.5 $\pm$ 2.3 & 34.5 $\pm$ 2.4\\
\hline

    \end{tabular}
    }
    \caption{
Thumos Class-Agnostic Detection Results.
 Performance of class-agnostic detectors with different input features. mAP@0.5IoUs is the mean Average 
Precision at 0.5 Intersection over Union. mAP@avg is the average of
mAP@0.3:0.1:0.7. AR@10 is Average Recall at 10. Aggregates are means 
$\pm$ standard error across splits.}
    \label{tab:thumos_proposals_supp}
\end{table}

\begin{table}[t]
    \centering
    \scalebox{0.8}{
    \begin{tabular}{|c|c|c|c|c|c|c|}
    \hline
    Detection Model & Classification Model & Split(s) & mAP@0.5 & mAP@0.75 & mAP@avg \\
    \hline

Clip b/32 & Clip b/32 & 75/25 & 32.6 $\pm$ 0.6 & 18.7 $\pm$ 0.5 & 19.4 $\pm$ 0.4\\
Clip b/32 & Clip b/32 & Intermediate & 34.7 & 20.4 & 21.0\\
Clip b/32 & Clip b/32 & Smart & 36.7 & 22.4 & 22.6\\
Clip b/32 & Clip b/32 & 50/50 & 30.1 $\pm$ 0.8 & 16.5 $\pm$ 0.5 & 17.3 $\pm$ 0.6\\
\hline
Clip b/16 & Clip b/16 & 75/25 & 36.3 $\pm$ 0.7 & 20.7 $\pm$ 0.6 & 21.4 $\pm$ 0.5\\
Clip b/16 & Clip b/16 & Intermediate & 38.6 & 22.7 & 23.4\\
Clip b/16 & Clip b/16 & Smart & 38.1 & 22.6 & 23.4\\
\hline

    \end{tabular}
    }
    \caption {
ActivityNet Open Vocabulary Detection Results. performance of open-vocabulary temporal action detection for different combinations of class-agnostic detection features and open-vocabulary classification features. mAP@0.5IoUs is the mean Average 
Precision at 0.5 Intersection over Union. mAP@avg is the average of
mAP@0.50:0.05:0.95. AR@10 is Average Recall at 10. Aggregates are means 
$\pm$ standard error across splits.}
      \label{tab:anet_e2e_supp}
\end{table}

\begin{table}[t]
    \centering
    \scalebox{0.6}{
    \begin{tabular}{|c|c|c|c|c|c|c|c|c|}
    \hline
    Detection Model & Classification Model & Split(s) & mAP@0.3 & mAP@0.4 & mAP@0.5 & mAP@0.6 & mAP@0.7 & mAP@avg  \\
    \hline

Clip b/32 & Clip b/32 & 75/25 & 21.4 $\pm$ 2.0 & 16.4 $\pm$ 1.8 & 11.8 $\pm$ 1.6 & 7.0 $\pm$ 1.2 & 3.5 $\pm$ 0.8 & 12.0 $\pm$ 1.4\\
Clip b/32 & Clip b/32 & 50/50 & 15.4 $\pm$ 3.3 & 11.0 $\pm$ 2.4 & 7.7 $\pm$ 1.9 & 4.3 $\pm$ 1.3 & 1.9 $\pm$ 0.7 & 8.0 $\pm$ 1.8\\
\hline
Clip b/16 & Clip b/16 & 75/25 & 21.8 $\pm$ 1.8 & 17.3 $\pm$ 1.7 & 13.2 $\pm$ 1.6 & 8.8 $\pm$ 1.6 & 3.7 $\pm$ 0.8 & 12.9 $\pm$ 1.3\\
Clip b/16 & Clip b/16 & 50/50 & 18.0 $\pm$ 1.5 & 13.2 $\pm$ 1.1 & 8.9 $\pm$ 0.9 & 5.1 $\pm$ 0.9 & 2.2 $\pm$ 0.7 & 9.5 $\pm$ 0.8\\
\hline
Clip l/14 & Clip l/14 & 75/25 & 28.6 $\pm$ 2.5 & 21.7 $\pm$ 2.0 & 15.4 $\pm$ 1.6 & 9.2 $\pm$ 1.2 & 4.2 $\pm$ 0.8 & 15.8 $\pm$ 1.6\\
Clip l/14 & Clip l/14 & 50/50 & 21.0 $\pm$ 3.7 & 14.8 $\pm$ 2.4 & 9.8 $\pm$ 2.0 & 5.0 $\pm$ 1.1 & 2.0 $\pm$ 0.3 & 10.5 $\pm$ 1.9\\
\hline
Align & Align & 75/25 & 15.1 $\pm$ 2.3 & 12.2 $\pm$ 2.0 & 8.9 $\pm$ 1.8 & 5.6 $\pm$ 1.7 & 1.7 $\pm$ 0.4 & 8.7 $\pm$ 1.5\\
\hline

    \end{tabular}
    }
    \caption {
Thumos Open Vocabulary Detection Results. performance
 of open-vocabulary temporal action detection for different combinations of 
 class-agnostic detection features and open-vocabulary classification features. mAP@0.5IoUs is the mean Average 
Precision at 0.5 Intersection over Union. mAP@avg is the average of
mAP@0.3:0.1:0.7. Aggregates are means $\pm$ standard error across splits.}
      \label{tab:thumos_e2e_supp}
\end{table}

\begin{table}[t]
    \centering
    \scalebox{0.45}{
    \begin{tabular}{|c|c|c|c|c|c|c|c|}
    \hline
    Model & mAP@0.5 & mAP@0.75 & mAP@0.95 & mAP@avg & AR@10 & AR@50 & AR@100 \\
    \hline

\multicolumn{8}{|c|}{Clip b/32}\\
\hline
$+$I3D\_FLOW - Intermediate & 50.2 & 28.5 & 3.5 & 29.2 & 43.4 & 48.9 & 48.9\\
$+$I3D\_FLOW - Smart & 53.0 & 29.7 & 3.0 & 30.6 & 44.2 & 47.2 & 47.2\\
$+$I3D\_RGB - Intermediate & 48.8 & 28.2 & 3.6 & 28.8 & 43.1 & 48.2 & 48.2\\
$+$I3D\_RGB - Smart & 51.2 & 30.5 & 3.7 & 30.7 & 44.6 & 47.3 & 47.3\\
$+$YAMNET - Intermediate & 48.3 & 27.5 & 4.5 & 28.5 & 42.4 & 50.1 & 50.1\\
$+$YAMNET - Smart & 50.3 & 29.4 & 5.4 & 30.3 & 44.6 & 49.2 & 49.2\\
\hline

    \end{tabular}
    }
    \caption{
ActivityNet Class-Agnostic Detection Results for Feature Ensembles on a Subset of Splits.
 Performance of class-agnostic detectors with different input feature ensembles. mAP@0.5IoUs is the mean Average 
Precision at 0.5 Intersection over Union. mAP@avg is the average of
mAP@0.50:0.05:0.95. AR@10 is Average Recall at 10. Aggregates are means 
$\pm$ standard error across splits.}
    \label{tab:anet_merged_subset_proposals_supp}
\end{table}

\begin{table}[t]
    \centering
    \scalebox{0.6}{
    \begin{tabular}{|c|c|c|c|c|c|c|c|}
    \hline
    Model & mAP@0.5 & mAP@0.75 & mAP@0.95 & mAP@avg & AR@10 & AR@50 & AR@100 \\
    \hline

\multicolumn{8}{|c|}{Clip b/32}\\
\hline
$+$I3D\_FLOW - 75/25 & 49.3 $\pm$ 0.7 & 27.3 $\pm$ 0.6 & 2.7 $\pm$ 0.2 & 28.2 $\pm$ 0.5 & 42.0 $\pm$ 0.5 & 47.6 $\pm$ 0.3 & 47.6 $\pm$ 0.3\\
$+$I3D\_FLOW - Smart & 53.0 & 29.7 & 3.0 & 30.6 & 44.2 & 47.2 & 47.2\\
$+$I3D\_FLOW - 50/50 & 47.5 $\pm$ 0.9 & 25.6 $\pm$ 1.2 & 2.7 $\pm$ 0.5 & 26.8 $\pm$ 1.1 & 40.2 $\pm$ 1.4 & 45.1 $\pm$ 1.7 & 45.1 $\pm$ 1.7\\
\hline
$+$I3D\_FLOW$+$VGGISH - 75/25 & 49.1 $\pm$ 0.6 & 27.3 $\pm$ 0.6 & 3.0 $\pm$ 0.2 & 28.2 $\pm$ 0.5 & 42.7 $\pm$ 0.4 & 47.4 $\pm$ 0.4 & 47.4 $\pm$ 0.4\\
$+$I3D\_FLOW$+$VGGISH - Smart & 52.9 & 30.6 & 4.0 & 31.5 & 44.7 & 49.3 & 49.3\\
$+$I3D\_FLOW$+$VGGISH - 50/50 & 46.5 $\pm$ 0.8 & 25.8 $\pm$ 0.7 & 2.7 $\pm$ 0.2 & 26.6 $\pm$ 0.5 & 40.2 $\pm$ 0.6 & 45.9 $\pm$ 0.7 & 45.9 $\pm$ 0.7\\
\hline

    \end{tabular}
    }
    \caption{
ActivityNet Class-Agnostic Detection Results for Feature Ensembles.
 Performance of class-agnostic detectors with different input feature ensembles. mAP@0.5IoUs is the mean Average 
Precision at 0.5 Intersection over Union. mAP@avg is the average of
mAP@0.50:0.05:0.95. AR@10 is Average Recall at 10. Aggregates are means 
$\pm$ standard error across splits.}
    \label{tab:anet_merged_proposals_supp}
\end{table}

\begin{table}[t]
    \centering
    \scalebox{0.5}{
    \begin{tabular}{|c|c|c|c|c|c|c|c|c|c|}
    \hline
    Model & mAP@0.3 & mAP@0.4 & mAP@0.5 & mAP@0.6 & mAP@0.7 & mAP@avg & AR@10 & AR@50 & AR@100 \\
    \hline

\multicolumn{10}{|c|}{Clip b/32}\\
\hline
$+$I3D\_FLOW - 75/25 & 43.3 $\pm$ 4.8 & 35.9 $\pm$ 4.3 & 27.6 $\pm$ 3.8 & 19.4 $\pm$ 3.5 & 12.2 $\pm$ 2.9 & 27.7 $\pm$ 3.7 & 11.9 $\pm$ 1.9 & 28.3 $\pm$ 2.4 & 35.9 $\pm$ 2.5\\
$+$I3D\_FLOW - 50/50 & 45.9 $\pm$ 2.6 & 35.8 $\pm$ 3.2 & 25.4 $\pm$ 4.0 & 15.4 $\pm$ 3.9 & 8.2 $\pm$ 3.3 & 26.1 $\pm$ 3.2 & 10.5 $\pm$ 2.5 & 26.1 $\pm$ 4.2 & 31.6 $\pm$ 4.4\\
\hline
$+$I3D\_FLOW$+$VGGISH - 75/25 & 48.3 $\pm$ 4.2 & 40.3 $\pm$ 4.3 & 30.8 $\pm$ 4.4 & 21.2 $\pm$ 4.2 & 13.4 $\pm$ 3.9 & 30.8 $\pm$ 4.1 & 13.6 $\pm$ 2.0 & 29.9 $\pm$ 2.5 & 35.4 $\pm$ 2.4\\
$+$I3D\_FLOW$+$VGGISH - 50/50 & 47.7 $\pm$ 2.7 & 36.1 $\pm$ 3.5 & 24.5 $\pm$ 3.8 & 14.3 $\pm$ 3.7 & 7.3 $\pm$ 3.0 & 26.0 $\pm$ 3.1 & 10.4 $\pm$ 2.0 & 23.4 $\pm$ 3.2 & 28.5 $\pm$ 3.0\\
\hline
$+$I3D\_FLOW$+$I3D\_RGB$+$VGGISH - 75/25 & 53.1 $\pm$ 3.8 & 45.0 $\pm$ 3.9 & 35.1 $\pm$ 4.1 & 24.5 $\pm$ 4.2 & 15.4 $\pm$ 4.0 & 34.6 $\pm$ 3.9 & 13.8 $\pm$ 1.8 & 30.9 $\pm$ 2.4 & 36.6 $\pm$ 2.4\\
$+$I3D\_FLOW$+$I3D\_RGB$+$VGGISH - 50/50 & 48.1 $\pm$ 4.2 & 37.8 $\pm$ 4.3 & 26.3 $\pm$ 3.9 & 15.6 $\pm$ 3.7 & 7.8 $\pm$ 3.1 & 27.1 $\pm$ 3.6 & 10.6 $\pm$ 1.9 & 25.4 $\pm$ 2.8 & 30.8 $\pm$ 2.3\\
\hline

    \end{tabular}
    }
    \caption{
Thumos Class-Agnostic Detection Results for Feature Ensembles.
 Performance of class-agnostic detectors with different input feature ensembles. mAP@0.5IoUs is the mean Average 
Precision at 0.5 Intersection over Union. mAP@avg is the average of
mAP@0.3:0.1:0.7. AR@10 is Average Recall at 10. Aggregates are means 
$\pm$ standard error across splits.}
    \label{tab:thumos_merged_proposals_supp}
\end{table}

\begin{table}[t]
    \centering
    \scalebox{0.49}{
    \begin{tabular}{|c|c|c|c|c|}
    \hline
    Added Features & Split(s) & mAP@0.5 & mAP@0.75 & mAP@avg \\
    \hline

\multicolumn{5}{|c|}{Clip b/32}\\
\hline
 & 75/25 & 32.6 $\pm$ 0.6 & 18.7 $\pm$ 0.5 & 19.4 $\pm$ 0.4\\
 & Intermediate & 34.7 & 20.4 & 21.0\\
 & Smart & 36.7 & 22.4 & 22.6\\
 & 50/50 & 30.1 $\pm$ 0.8 & 16.5 $\pm$ 0.5 & 17.3 $\pm$ 0.6\\
\hline
$+$I3D\_FLOW & 75/25 & 33.0 $\pm$ 0.6 & 19.6 $\pm$ 0.6 & 19.9 $\pm$ 0.5\\
$+$I3D\_FLOW & Intermediate & 35.2 & 21.6 & 21.8\\
\hline

    \end{tabular}
    }
    \caption {
ActivityNet Open Vocabulary Detection Results with Feature Ensembles.
performance of open-vocabulary temporal action detection for different 
combinations of class-agnostic detection features and and the listed 
open-vocabulary classification features. mAP@0.5IoUs is the mean Average 
Precision at 0.5 Intersection over Union. mAP@avg is the average of
mAP@0.50:0.05:0.95. Aggregates are means $\pm$ standard error across splits.}
      \label{tab:anet_e2e_merged_supp}
\end{table}

\begin{table}[t]
    \centering
    \scalebox{0.6}{
    \begin{tabular}{|c|c|c|c|c|c|c|c|}
    \hline
    Added Features & Split(s) & mAP@0.3 & mAP@0.4 & mAP@0.5 & mAP@0.6 & mAP@0.7 & mAP@avg  \\
    \hline
\multicolumn{8}{|c|}{Clip b/32}\\
\hline
 & 75/25 & 21.4 $\pm$ 2.0 & 16.4 $\pm$ 1.8 & 11.8 $\pm$ 1.6 & 7.0 $\pm$ 1.2 & 3.5 $\pm$ 0.8 & 12.0 $\pm$ 1.4\\
 & 50/50 & 15.4 $\pm$ 3.3 & 11.0 $\pm$ 2.4 & 7.7 $\pm$ 1.9 & 4.3 $\pm$ 1.3 & 1.9 $\pm$ 0.7 & 8.0 $\pm$ 1.8\\
\hline
$+$I3D\_FLOW & 75/25 & 19.7 $\pm$ 2.8 & 16.0 $\pm$ 2.4 & 11.7 $\pm$ 1.8 & 7.0 $\pm$ 1.2 & 3.3 $\pm$ 0.7 & 11.5 $\pm$ 1.7\\
$+$I3D\_FLOW & 50/50 & 20.1 $\pm$ 2.6 & 15.3 $\pm$ 1.7 & 11.1 $\pm$ 1.5 & 6.3 $\pm$ 1.3 & 2.9 $\pm$ 1.0 & 11.1 $\pm$ 1.3\\
\hline
$+$I3D\_FLOW$+$VGGISH & 75/25 & 22.3 $\pm$ 2.5 & 17.5 $\pm$ 2.2 & 12.7 $\pm$ 1.7 & 7.5 $\pm$ 1.2 & 3.5 $\pm$ 0.7 & 12.7 $\pm$ 1.6\\
$+$I3D\_FLOW$+$VGGISH & 50/50 & 17.5 $\pm$ 5.2 & 13.0 $\pm$ 3.7 & 8.9 $\pm$ 2.5 & 4.8 $\pm$ 1.3 & 1.9 $\pm$ 0.4 & 9.2 $\pm$ 2.6\\
\hline

    \end{tabular}
    }
    \caption {
Thumos Open Vocabulary Detection Results with Feature Ensembles.
Performance of open-vocabulary temporal action detection for different 
combinations of class-agnostic detection features and and the listed 
open-vocabulary classification features. mAP@0.5IoUs is the mean Average 
Precision at 0.5 Intersection over Union. mAP@avg is the average of
mAP@0.3:0.1:0.7. Aggregates are means $\pm$ standard error across splits.}
      \label{tab:thumos_e2e_merged_supp}
\end{table}

\begin{table}[t]
    \centering
    \scalebox{0.45}{
    \begin{tabular}{|c|c|c|c|c|c|}
    \hline
    Intermediate & r1 & r2 & r3 & r4 & r5 \\
    \hline

Zumba & Drinking coffee & Doing kickboxing & Tango & Zumba & Drinking coffee\\
Putting on makeup & Playing polo & Snatch & Long jump & Tango & Doing kickboxing\\
Ironing clothes & Cricket & Clean and jerk & Using parallel bars & Putting on makeup & Playing polo\\
Discus throw & Bathing dog & Playing field hockey & Grooming horse & Long jump & Snatch\\
Doing karate & Preparing salad & Playing harmonica & Pole vault & Ironing clothes & Cricket\\
Chopping wood & Playing saxophone & Washing face & Using the pommel horse & Using parallel bars & Clean and jerk\\
Spinning & Javelin throw & Ping-pong & Volleyball & Discus throw & Bathing dog\\
\hline
    \end{tabular}
    }
    \caption{ActivityNet 75/25 random splits.
     The strings listed are the ActivityNet labels held out for evaluation.
      The first column is the Intermediate split mentioned in the main text.}
    \label{tab:anet_7525_splits_supp}
\end{table}

\begin{table}[t]
    \centering
    \scalebox{0.45}{
    \begin{tabular}{|c|c|c|c|c|c|}
    \hline
    r0 & r1 & r2 & r3 \\
    \hline

Zumba & Drinking coffee & Drinking coffee & Arm wrestling\\
Tango & Doing kickboxing & Zumba & Assembling bicycle\\
Putting on makeup & Playing polo & Doing kickboxing & BMX\\
Long jump & Snatch & Tango & Baking cookies\\
Ironing clothes & Cricket & Playing polo & Baton twirling\\
Using parallel bars & Clean and jerk & Putting on makeup & Beach soccer\\
Discus throw & Bathing dog & Snatch & Beer pong\\
\hline
    \end{tabular}
    }
    \caption{ActivityNet 50/50 random splits.
     The strings listed are the ActivityNet labels held out for evaluation.}
    \label{tab:anet_5050_splits_supp}
\end{table}

\begin{table}[t]
    \centering
    \scalebox{0.7}{
    \begin{tabular}{|c|c|c|c|c|c|}
    \hline
    r0 & r1 & r2 & r3 & r4 & r5 \\
    \hline

BaseballPitch & BasketballDunk & Billiards & CleanAndJerk & BasketballDunk & BaseballPitch\\
CliffDiving & CricketBowling & CricketShot & Diving & CleanAndJerk & Billiards\\
FrisbeeCatch & GolfSwing & HammerThrow & HighJump & CricketBowling & CliffDiving\\
JavelinThrow & LongJump & PoleVault & Shotput & Diving & CricketShot\\
SoccerPenalty & TennisSwing & ThrowDiscus & VolleyballSpiking & GolfSwing & FrisbeeCatch\\
\hline
r6 & r7 & r8 & r9 & r10 & r11 \\
\hline
HighJump & HammerThrow & BaseballPitch & CricketBowling & HammerThrow & Shotput\\
LongJump & JavelinThrow & BasketballDunk & CricketShot & HighJump & SoccerPenalty\\
Shotput & PoleVault & Billiards & Diving & JavelinThrow & TennisSwing\\
TennisSwing & SoccerPenalty & CleanAndJerk & FrisbeeCatch & LongJump & ThrowDiscus\\
VolleyballSpiking & ThrowDiscus & CliffDiving & GolfSwing & PoleVault & VolleyballSpiking\\

\hline
    \end{tabular}
    }
    \caption{Thumos 75/25 random splits.
     The strings listed are the Thumos labels held out for evaluation.}
    \label{tab:thumos_7525_splits_supp}
\end{table}

\begin{table}[t]
    \centering
    \scalebox{0.7}{
    \begin{tabular}{|c|c|c|c|}
    \hline
    r0 & r1 & r2 & r3 \\
    \hline

BasketballDunk & BaseballPitch & BaseballPitch & HammerThrow\\
CleanAndJerk & Billiards & BasketballDunk & HighJump\\
CricketBowling & CliffDiving & Billiards & JavelinThrow\\
Diving & CricketShot & CleanAndJerk & LongJump\\
GolfSwing & FrisbeeCatch & CliffDiving & PoleVault\\
HighJump & HammerThrow & CricketBowling & Shotput\\
LongJump & JavelinThrow & CricketShot & SoccerPenalty\\
Shotput & PoleVault & Diving & TennisSwing\\
TennisSwing & SoccerPenalty & FrisbeeCatch & ThrowDiscus\\
VolleyballSpiking & ThrowDiscus & GolfSwing & VolleyballSpiking\\

\hline
    \end{tabular}
    }
    \caption{Thumos 50/50 random splits.
     The strings listed are the Thumos labels held out for evaluation.}
    \label{tab:thumos_5050_splits_supp}
\end{table}

\clearpage

\end{document}